\documentclass{article}

\usepackage{microtype}
\usepackage{graphicx}
\usepackage{subfigure}
\usepackage{booktabs} 

\usepackage{hyperref}
\usepackage[table,xcdraw]{xcolor}

\usepackage[accepted]{icml2023}

\usepackage{amsmath}
\usepackage{amssymb}
\usepackage{mathtools}
\usepackage{amsthm}

\usepackage[capitalize,noabbrev]{cleveref}

\theoremstyle{plain}

\theoremstyle{definition}

\theoremstyle{remark}

\usepackage[textsize=tiny]{todonotes}

\icmltitlerunning{Heterogeneous Image GNN: Graph-Conditioned Diffusion for Image Synthesis}

\makeatletter
\renewcommand{\@copyrightspace}{}
\renewcommand{\Notice@String}{}
\renewcommand{\ICML@appearing}{}
\makeatother

\begin{document}

\twocolumn[
\icmltitle{Heterogeneous Image GNN: Graph-Conditioned Diffusion for Image Synthesis}




\begin{icmlauthorlist}
\icmlauthor{Rupert Menneer\textsuperscript{*}}{cam}
\icmlauthor{Christos Margadji}{cam}
\icmlauthor{Sebastian W. Pattinson}{cam}
\end{icmlauthorlist}

\icmlaffiliation{cam}{Department of Engineering, University of Cambridge, Cambridge, United Kingdom}

\icmlcorrespondingauthor{Rupert Menneer}{rfsm2@cam.ac.uk}
\icmlcorrespondingauthor{Sebastian W. Pattinson}{swp29@cam.ac.uk}



\icmlkeywords{Machine Learning, ICML}

\vskip 0.3in
]



\printAffiliationsAndNotice{} 

\begin{abstract}
We introduce a novel method for conditioning diffusion-based image synthesis models with heterogeneous graph data. Existing approaches typically incorporate conditioning variables directly into model architectures, either through cross-attention layers that attend to text latents or image concatenation that spatially restrict generation. However, these methods struggle to handle complex scenarios involving diverse, relational conditioning variables, which are more naturally represented as unstructured graphs. This paper presents Heterogeneous Image Graphs (HIG), a novel representation that models conditioning variables and target images as two interconnected graphs, enabling efficient handling of variable-length conditioning inputs and their relationships. We also propose a magnitude-preserving GNN that integrates the HIG into the existing EDM2 diffusion model using a ControlNet approach. Our approach improves upon the SOTA on a variety of conditioning inputs for the COCO-stuff and Visual Genome datasets, and showcases the ability to condition on graph attributes and relationships represented by edges in the HIG.  
\end{abstract}

\section{Introduction}
\label{sec:intro}

\begin{figure}[t]
    \centering
\includegraphics[width=1\linewidth]{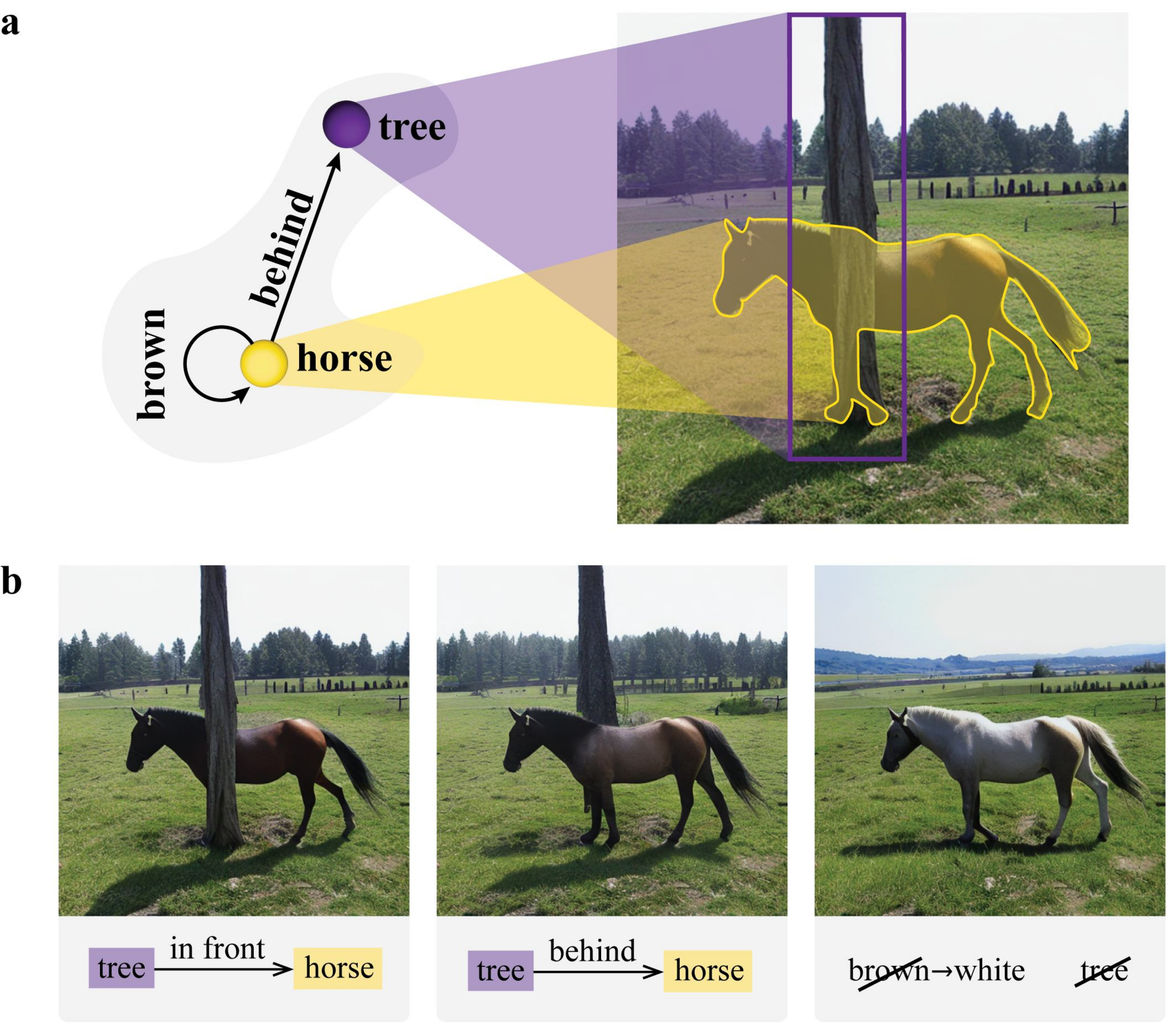}
    \vspace{-20bp}
    \caption{
\textbf{(a)} Our representation enables flexible conditioning for graph-to-image generation by modeling objects, attributes, and relationships as a graph. This integrates with a pixel grid, where pixels act as nodes connected to objects, defining spatial relationships. \textbf{(b)} Modifying conditions at the node and or edge level enables precise semantic control over the generation process.}
\label{fig:fig_1_intro}
\end{figure}


Conditional denoising diffusion models have become widely popular due to their proficiency for high-quality and controlled image synthesis \cite{saharia_image_2021, saharia_photorealistic_2022, karras_analyzing_2024, li_controlnet_2024, zhang_adding_2023, ramesh_hierarchical_2022}. Diffusion models transform pure noise into images through repeated application of denoising steps via a learned denoiser \cite{ho_denoising_2020, song_score-based_2021}. This sampling process can be modelled through differential equations, where each denoising step can be seen through the lens of score matching \cite{song_score-based_2021, hyvarinen_estimation_2005}. Typically, the image processing architecture is a U-Net model with intermediate self-attention layers \cite{ho_denoising_2020, saharia_image_2021, karras_analyzing_2024}. Controlling generation is achieved through two routes: explicit conditioning and guidance during sampling \cite{saharia_image_2021, karras_guiding_2024, ho_classifier-free_2022}. Explicit conditioning provides conditioning information directly to the model, for example text prompts or low-resolution images \cite{saharia_photorealistic_2022}. However, real-world applications such as image-editing, weather-modelling and additive manufacturing demand the generation of images conditioned on variable-length, and diversely structured data i.e. graph-to-image – a challenge that remains largely unaddressed.


Concurrently, the field of geometric deep learning has gained  traction by extending deep learning techniques to non-regular data-spaces such as graphs i.e.  graph neural networks (GNNs) \cite{kipf_semi-supervised_2017, hamilton_inductive_2018, velickovic_graph_2018, gilmer_neural_2017}. In particular, this has allowed advancements in weather modelling where data on graph nodes represent spatial relationships between geographical locations \cite{lam_graphcast_2022}. GNNs have also been applied in the image domain breaking the rigidity of traditional grid-based techniques \cite{tian_image_nodate, han_vision_2022}. Moreover, they have been explored in image synthesis with diffusion models, most notably for image generation from scene graphs that are designed to contain coarse structures of the scene images \cite{farshad_scenegenie_2023, johnson_image_2018, yang_diffusion-based_2022}. By processing a scene graph with a graph encoder, they produce local and global latents that aid in accurate generation. However, these fall short of a general-purpose approach that can seamlessly interweave heterogeneous, sparse, and variable length graphs whilst retaining the power of existing image models.

In this paper we introduce Heterogenous Image Graphs (HIG), a novel representation for conditional image generation that leverages the power of diffusion models in conjunction with GNNs. This representation, illustrated in Figure \ref{fig:fig_1_intro}, allows an image to switch between a standard image representation and a set of image patch or pixel nodes. We then consider a second `conditioning' graph and its relationships to both the image nodes and itself. By switching between these representations, we allow the generated image to be processed by tried-and-true architectures such as the U-Net \cite{ronneberger_u-net_2015, karras_analyzing_2024}, whereas the HIG representation can be processed intermediately by a GNN. This allows complex conditioning variables and relationships to be processed directly within the model architecture. We demonstrate that this method effectively conditions both semantic and relational information. To achieve this, we apply the magnitude-preserving formulation set out in EDM2 \cite{karras_analyzing_2024} to a custom graph convolution operator, and show it preserves magnitudes under certain conditions required of the graph data.


This approach proves to be an effective and adaptable method, seamlessly integrating diverse conditions from multiple datasets. We evaluate our trained model on a variety of tasks and show SOTA performance in multiple metrics. For the Visual Genome layout-to-Image task, our approach improves the previous record FID from 15.63 to 8.79, and  15.61 to 11.41 for the COCO-stuff mask-to-Image task, at higher resolution compared to previous work.

Our main contributions are:

\vspace{-4mm} 
\begin{itemize}
    \item The HIG representation, demonstrating its effectiveness in fine-grained image control through graph-based attributes and relationships. 
    \item SOTA results on layout-to-image and mask-to-image tasks (COCO-Stuff, VG) at 512×512 resolution.
\end{itemize}

\section{Related Work}
\label{sec:related_work}

Since the focus of this work is the effective and flexible conditioning of image synthesis with conditioning graphs, and we do not contribute to the theory behind diffusion, we direct readers to prior works \cite{sohl-dickstein_deep_2015, song_score-based_2021, ho_denoising_2020, karras_elucidating_2022} for the mathematical preliminaries. Readers should be aware of Karras et al. \cite{karras_analyzing_2024} which forms the basis of this work in regards to diffusion architecture design. A brief summary of conditional diffusion in this context is provided below for reader convenience.

\textbf{Conditional Diffusion.} Conditional diffusion extends the standard diffusion framework by introducing conditional variables into the generative process, allowing for control over the output. Instead of modeling the data distribution \( p(\mathbf{x}; \sigma(t)) \) with time-dependent noise level $\sigma$, conditional diffusion focuses on generating data given conditioning variables \( \mathbf{c} \), resulting in \( p(\mathbf{x} | \mathbf{c}; \sigma(t)) \). Both approaches can be described through SDE or ODE formulations. The probability ODE formulation from Karras et al. \cite{karras_elucidating_2022} describes the process both forward and backward in time:

\begin{equation}
    dx = -\dot{\sigma}(t)\sigma(t) \nabla_x \log p(x| \mathbf{c}; \sigma(t)) dt,
    \label{eq:karras_1}
\end{equation}

where \(\dot{\sigma}(t)\) is the time derivative, and \( \nabla_{\mathbf{x}} \log p_t(\mathbf{x} | \mathbf{c}; \sigma(t)) \) represents the conditional score function that depends on the noise level and conditioning variables $\mathbf{c}$. This score function can be approximated through an L2 denoising objective \cite{karras_elucidating_2022, song_score-based_2021}. This approach allows for fast deterministic and higher-order sampling which has been shown to be highly effective \cite{karras_elucidating_2022}.

Typically, conditioning is done by explicitly providing conditioning signals such as text embeddings, low-resolution images, or other signals to the model \cite{saharia_photorealistic_2022, saharia_image_2021, li_controlnet_2024, dhariwal_diffusion_2021}. This can be done through simple concatenation \cite{saharia_image_2021}, adaptive normalisation layers \cite{peebles_scalable_2023}, or cross-attention mechanisms \cite{saharia_photorealistic_2022}. For example, in the \textit{Imagen} framework of Saharia et al. \cite{saharia_photorealistic_2022}, conditional superresolution is achieved by conditioning text latents from a pre-trained language model via cross-attention, and conditioning on low-resolution images via concatenation. Another effective conditioning approach was proposed in the \textit{ControlNet} framework of Zhang et al. \cite{zhang_adding_2023}, where diverse spatial-conditional generation was achieved by incorporating an additional network alongside a large, pre-trained text-to-image diffusion model with frozen parameters. Conditioning signals are incorporated by first transforming the conditions to latent image space and passing them as input. Finally, in the EDM2 model \cite{karras_analyzing_2024} - the focus of this work - they generate images conditioned on simple class labels via simple multiplication with a learned embedding and a zero-initialised gain parameter. Notably, this work also deeply analyses the training dynamics of diffusion models and highlights the importance of standardising weight magnitudes explicitly by design, an approach we adopt.

\textbf{Diffusion with Conditioning Graphs.} Graphs have previously been used to condition models for image synthesis. In particular, one popular task is generating images that adhere to a scene graph that represents the image \cite{johnson_image_2018, yang_diffusion-based_2022, dhamo_semantic_2020, farshad_scenegenie_2023, mittal_interactive_2019}. Typically, scene graphs are first converted to an intermediate layout (i.e. a latent image) before being processed by a secondary network such as diffusion models. Farshad et al. \cite{farshad_scenegenie_2023} process graphs with a GCN to predict object embeddings that control the network via sampling guidance. Yang et al. \cite{yang_diffusion-based_2022} pre-train a masked auto-encoder on scene graph triplets to generate local and global embeddings to condition the diffusion model. Modifying scene graphs have also been used as a way for users to interactively control image synthesis \cite{mittal_interactive_2019, dhamo_semantic_2020}. 

\textbf{Graphs for Image Processing.} GNNs have been used in the computer vision domain for various applications including image processing where images are viewed directly as graphs \cite{tarasiewicz_graph_2021, han_vision_2022, tian_image_nodate, krzywda_graph_2022, defferrard_convolutional_2017, liu_cnn-enhanced_2021, wan_multi-scale_2019}. For example, Han et al. \cite{han_vision_2022} demonstrate GNNs to be effective for image classification tasks by splitting images into nodes and connecting them via nearest neighbours. Tian et al. \cite{tian_image_nodate} use GNNs for super-resolution by leveraging variable node degree to focus on high-frequency areas. Scene graph generation typically processes images with traditional CNN architectures, and decodes them with GNNs \cite{xu_scene_2017, yang_graph_2018}. While prior work demonstrates that graph-based models are flexible and powerful for image processing, their potential to directly use them to condition image diffusion models remains underexplored.

\section{Method}
\label{sec:method}

\subsection{Weaknesses of Previous Conditioning Methods}

The most popular form of latent image conditioning typically converts conditioning signals to images, before processing them with typical image processing models. While this approach is powerful, it exhibits limitations in handling complex image synthesis tasks, particularly when incorporating heterogeneous or sparse input conditions. Some approaches, such as \textit{LayoutDiffusion} \cite{zheng_layoutdiffusion_2024}, tackle this with custom attention modules that attend to bounding boxes with learned positional embeddings. However, these approaches neglect to include multiple modalities and the relationships between them, which overlooks nuanced interactions between conditioning signals i.e. disambiguating spatial ordering between overlapping boxes. 



Previous conditional diffusion research that utilise graph data opt for complex multi-stage training procedures such as masked contrastive pre-training using graph triplets \cite{yang_diffusion-based_2022}. This is not only time-consuming, but also fails to exploit potential benefits of training an end-to-end system that integrates graph data directly into image processing. 

We tackle these problems by representing images and their conditioning signals as a single graph, which is processed by a bespoke GNN architecture. This allows repeated interactions between conditioning signals and the image throughout the synthesis process, enabling more flexible and dynamic representations that account for both the current image features and interactions between conditioning signals. By maintaining separate pathways for distinct input types, our approach supports heterogeneous and sparse conditioning, leading to better generalisation, finer control, and more precise manipulation of generated images. This simple yet powerful method can be easily integrated into a wide range of existing vision models.

\begin{figure}
    \centering    \includegraphics[width=1\linewidth]{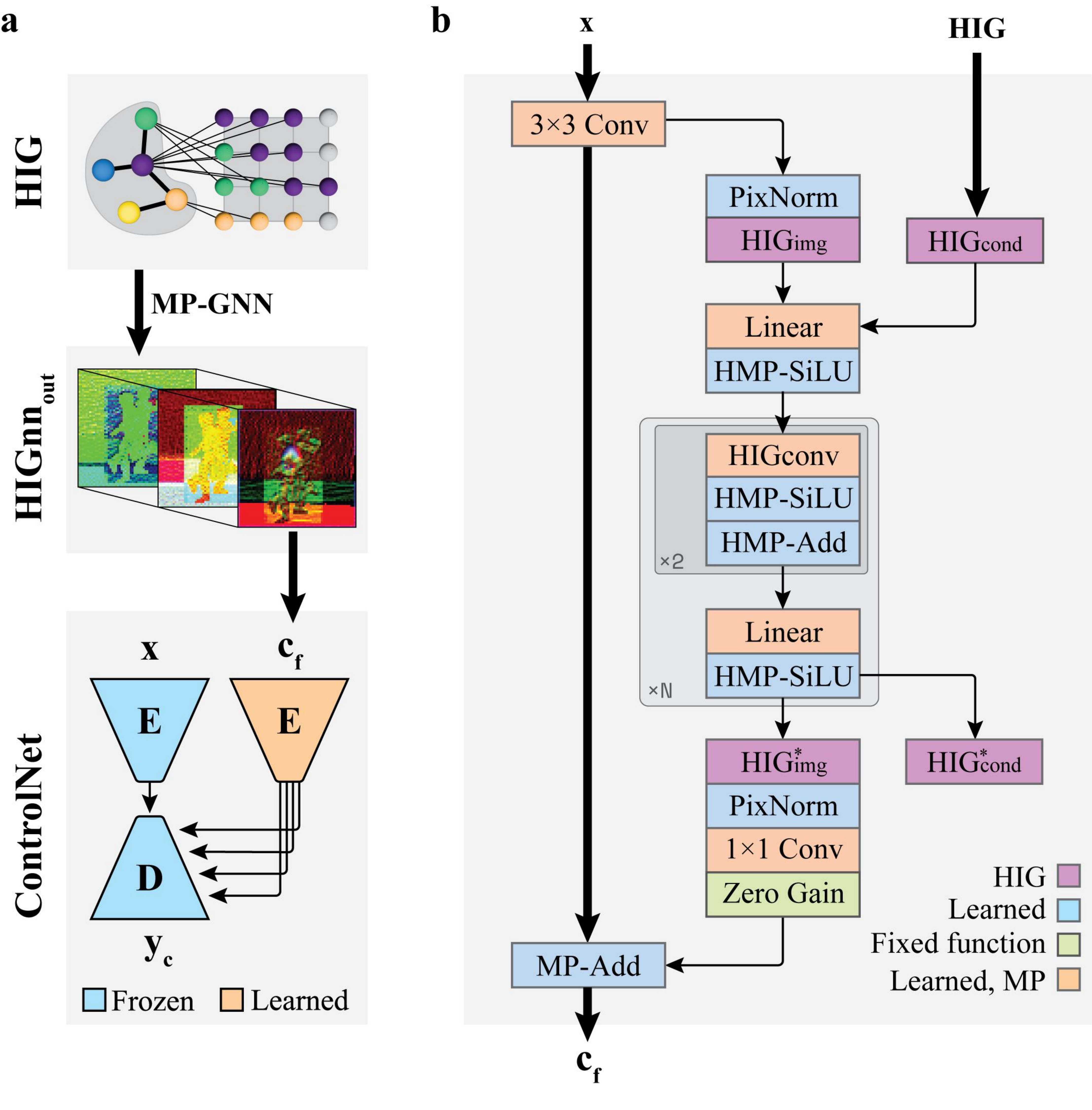}
\vspace{-20pt}
    \caption{(\textbf{a}) Overview of the proposed architecture. The HIG is encoded into a latent representation through a MP-GNN which is then used as a condition $c_f$ in a ControlNet. (\textbf{b}) Details of the MP-GNN module. Note: HMP is shorthand for heterogenous magnitude preserving operations applied across all nodes.}
    \label{fig:architecture}
\end{figure}

\subsection{Heterogeneous Image Graphs}

To improve on previous approaches we develop a new approach to condition images via the HIG representation. In this manner, we fully exploit variable-length and heterogeneous conditions to aid in image synthesis.

\textbf{Image Graphs.} When faced with the challenge of conditioning images with graphs we first convert images into representations amenable for graph processing. We reshape image features into image nodes pixel-wise in line with other works \cite{liu_cnn-enhanced_2021, han_vision_2022}. In practice, these nodes represent more than a single pixel, for example a latent image patch. This can be due to performing latent image diffusion \cite{rombach_high-resolution_2022, podell_sdxl_2023} where images are first pre-compressed to latent images, or due to prior processing by the image processing model. In contrast to other works \cite{tian_image_nodate, han_vision_2022, tarasiewicz_graph_2021}, we decide to leave image nodes unconnected; this loosely decouples image conditioning from processing. Image nodes are conditioned and later converted back into an image representation, allowing existing architectures to handle processing. Connecting image nodes in a locally dense fashion gains little benefit over highly optimised $3 \times 3$ convolutional operations. Formally, image nodes exist in a discrete space \( f : \mathbb{Z}^2 \to \mathbb{R}^C \). For an image of size \(M \times N\), we define \( f(i, j) \) where \( i, j \in \mathbb{Z} \) and \( 0 \leq i < M \), \( 0 \leq j < N \).

\textbf{Conditioning Graphs}. Conditioning graphs consist of nodes and edges, where each node has features defined as $ g : \mathcal{V} \to \mathbb{R}^F$, where $\mathcal{V}$ represents the set of nodes and $\mathbb{R}^F$ the feature space. Nodes may have spatial ties to the image domain, which we materialise via edges linking image and conditioning nodes. We use conditioning nodes to indicate semantics within the scene, for instance, a node may represent an object (e.g., a \textit{person}). Whereas we utilise different edge types to represent both spatial, abstract relationships and additional semantics. For instance, an edge between two object nodes may encode interactions or attributes (e.g., a person \textit{wearing} a {\textit{yellow}} hat). The graph structure reflects real-world data: often sparse and heterogeneous. We therefore construct graphs on a per task-basis to best leverage the available data and its dependencies.
Formally, each edge \( e \in \mathcal{E} \) connects two nodes \( (v_i, v_j) \in \mathcal{V} \times \mathcal{V} \) and represents a relationship between them. Edges represent any dependency, allowing for abstract relationships to be included.

\textbf{Connecting Image and Conditioning Nodes.} With image and conditioning nodes defined, we are close to the complete HIG representation. To enable conditioning between the image and conditioning graphs, we must construct edges between the two. These connections are determined on a per-task basis, depending on the available data, with explicit choices described in Section 4. However, when spatial information is available i.e. segmentation masks or bounding boxes, it enables direct connections between the image graph and the conditioning graph. Specifically, edges are created between image nodes relevant to spatial conditionings (i.e. pixels within the bounding box) and conditioning nodes representing the corresponding semantic class (i.e. class label). This linkage facilitates information flow across the graphs, integrating pixel-level details with higher-level semantic representations. 


\subsection{Model Architecture}

To be compatible with the EDM2 U-Net architecture \footnote{\href{https://github.com/NVlabs/edm2}{https://github.com/NVlabs/edm2}}, we propose the addition of a magnitude-preserving \textit{Heterogenous Image Graph Neural Network} (HIGnn) as the conditioning network to be used in a ControlNet strategy.

\textbf{HIGnn.} The general architecture of the HIG conditioning block requires two primary capabilities: representation switching and HIG processing. To handle switching between image features and image nodes on the HIG we consider the update function $\mathcal{U}_{\text{i}\rightarrow\text{g}}$. This update functions reshapes image features $\mathbf{x_i} \in \mathbb{R}^{N \times C \times H \times W}$ into image nodes pixel wise $\mathbf{x_g} \in \mathbb{R}^{N\cdot H \cdot W \times C}$ and applies an optional projection to ensure correct dimensionality. For the current set of image pixels $\mathbf{x_i}$, we retrieve HIG image nodes $\mathbf{x_g}$ by
\begin{equation}
\mathbf{x_g} = \mathcal{U}_{\text{i}\rightarrow\text{g}}(\mathbf{x_i}) = \hat{W}R(\mathbf{x_i}),  
 \label{eq:HIG_update}
\end{equation}
where $R$ reshapes the image, and $\hat{W}$ is a learned projection with forced magnitude preservation from \cite{karras_analyzing_2024}. Refer to Appendix \ref{appendix:edm2_preliminaries} for greater detail into the mathematical preliminaries of \cite{karras_analyzing_2024}. We consider the reverse operation of converting from graph nodes to an image $\mathcal{U}_{\text{g}\rightarrow\text{i}}$ in a similiar fashion. 

Once we have the HIG updated with current image nodes we can process it with a GNN. We identify several areas where magnitudes can grow and address them each in turn. In practice many varieties of heterogenous message passing GNN could be used, we create our own magnitude preserving graph convolutional operator similiar to Hamilton et al. \cite{hamilton_inductive_2018} for its simplicity and stability. The basic approach propagates information through two branches, a pseudo `skip-connection' applied to the current node, and a learned pooling operation of the local neighbourhood, and we add the ability to include edge information in the neighbourhood pooling. If edge attributes $\mathbf{a}_i$ are present we integrate them via magnitude preserving concatenation to the pooling branch. Formally, the HIGConv operator applied per meta-path to get updated node embeddings $\mathbf{x}_i'$ is defined as:
\begin{equation}
    \mathbf{x}_g' = \psi\left(\hat{W}^{\Phi}_1 \mathbf{x}_g 
    \underset{0 \text{ if } |\mathcal{N}^{\Phi}(i)| = 0}{\underbrace{+^\text{mp} \hat{W}^{\Phi}_2 \cdot \frac{1}{\sqrt{|\mathcal{N}^{\Phi}(i)|}} \sum_{j \in \mathcal{N}^{\Phi}(i)} [\mathbf{x}_j \|^\text{mp} \mathbf{a}_j]}}\right)    \label{eq:hignn_operator}
\end{equation}


where we choose $\psi$ to be magnitude preserving SiLU operator, and $+^\text{mp}$ the magnitude preserving sum (See Appendix \ref{appendix:edm2_preliminaries}), and both meta-path weights $\hat{W}^{\Phi}_1$ and $\hat{W}^{\Phi}_2$ have forced magnitude. $\mathcal{N}$ indicates the local node neighbourhood and is defined by the connectivity of graph. In order to achieve magnitude preservation we first assume all neighbourhood features to be of unit length, we then summate them scale them by the square root of the neighbourhood size ($\sqrt{|\mathcal{N}^{\Phi}|}$), see Appendix \ref{appendix:sum_random} for details. It is important to address unconnected or `zero-degree' nodes, in this case we ignore the right hand side of the equation, and only take the residual path. Note that simply setting the  neighbourhood to zero unintentionally changes the feature magnitudes when mp-sum is applied, since it assumes both vectors to be of unit length. Finally to combine information across meta-paths, we use the same method and sum across paths before normalising by the inverse square root of the number of incoming meta-paths ($|\Phi_i| = |\{\Phi_k \mid x_i \in \Phi_k\}|$)


\begin{equation}
\Tilde{\mathbf{x}}_g = \frac{1}{\sqrt{|\Phi_g|}} \sum_{\Phi \in \Phi_g} \mathbf{x}'_g,
\label{eq:meta_path}
\end{equation}

We verify that this approach is guaranteed to maintain magnitudes under certain conditions of the underlying graph data. In particular, for graph-data of sufficient size this approach holds for graphs which do not have identical features attached to the same node since this breaks the independence assumption. 


\textbf{EDM2 ControlNet Integration.} To integrate conditioning into a generative model, we adopt a strategy similar to ControlNet \cite{zhang_adding_2023}, i.e. a frozen EDM2 pre-trained model, with a trainable copy the encoder integrated with the conditioning HIGnn. Refer to Figure \ref{fig:architecture} for an overview of our proposed architecture, we employ 4 HIG blocks for our base model. The EDM2 checkpoints are only available for class-conditional generation of the 1000 ImageNet classes, yet we find them easy to adapt to our natural image datasets.  To facilitate this we unfreeze the embedding network. To integrate features we adopt $1\times1$ convolutions with a learnable zero-gain in a similar fashion to the original ControlNet, but we note that traditional summation may damage feature magnitudes. We find that naively integrating is harmful to training. Instead, we apply magnitude preserving summation, which, in contrast to the original ControlNet paper, directly alters the primary network features. This yields poor generative quality at step 0, but proves to be quick to train and to be best in practice.

In the trainable encoder we integrate our proposed HIGnn after the initial convolution block. We opt to keep the dimension of the GNN matched to that of the generative model. Finally, to generate samples we opt for the non-stochastic EDM2 sampler, and use the recent advancements in auto-guidance \cite{karras_guiding_2024}, we use our control model as the primary network, and use the unconditional XS ImageNet checkpoint released with EDM2 as the guidance network \cite{karras_analyzing_2024, karras_guiding_2024}. 

\section{Experiments}
\label{sec:experiments}



\subsection{Datasets}
We apply our method to two datasets: COCO-Stuff and Visual Genome both center cropped and resized to 512x512. We precompute latent images using the same autoencoder as in EDM2\footnote{\href{https://huggingface.co/stabilityai/sd-vae-ft-mse}{https://huggingface.co/stabilityai/sd-vae-ft-mse}}. We also employ random horizontal flipping and save our graph/image pairs to disk. For conditioning variables that involve natural language i.e. class labels, object relationships and image captions we opt to first encode them with a pretrained CLIP-ViT-Large\footnote{\href{https://huggingface.co/openai/clip-vit-large-patch14}{https://huggingface.co/openai/clip-vit-large-patch14}} model to get 768-d vectors for conditioning features. We precompute a vocabulary of class labels, attributes and relationships. This approach allows for generalised representations and for datasets to be fairly unfiltered when compared to previous work. Viewing all datasets through the lens of heterogeneous graphs allows us to jointly train on a variety of different conditioning signals, even when datasets contain different label modalities. Notably, these labels may be represented by vectors of variable length. 

\begin{table*}[t]
\label{table:metrics}
\centering
\caption{Results comparison between our method with guidance = 1.8 and LayoutDiffusion.}
\vspace{5bp}
\begin{tabular}{@{}lccccccc@{}}
\toprule & \multicolumn{4}{c}{\textbf{COCO-stuff}} & \multicolumn{3}{c}{\textbf{Visual Genome}} \\ 
\cmidrule(l){2-5} 
\cmidrule(l){6-8} 
\textbf{Methods} & \textbf{FID ↓} & \textbf{$\text{FID}_{\textit{DINOv2}}$ ↓} & \textbf{DS ↑} & \textbf{YOLOScore} ↑ & \textbf{FID ↓ }& \textbf{$\text{FID}_{\textit{DINOv2}}$ ↓}& \textbf{DS ↑} \\ \midrule

LayoutDiffusion (128$\times$128) &16.57&-&0.47±0.09&27.00&16.35&-&0.49±0.09\\
LayoutDiffusion (256$\times$256) &15.63&-&0.57±0.10&32.00&15.63&-&0.59±0.10\\
\textbf{HIG-Medium} (box only) & 11.63 &317.69&\textbf{0.67±0.10} &34.40& 8.99 & 367.89 & 0.68±0.10 \\
\textbf{HIG-Medium} & \textbf{11.42} & 256.93&0.59±0.11&\textbf{41.20}& N/A & N/A & N/A \\
\textbf{HIG-XXL} (box only) & 11.59 &  210.16 & 0.66±0.10 & 28.00 &\textbf{8.79}&213.80&0.64±0.10\\
\textbf{HIG-XXL} & 12.48 & \textbf{198.47}&0.56±0.12& 34.90&N/A&N/A& N/A \\
\bottomrule
\end{tabular}
\end{table*}

\textbf{COCO-stuff} consists of 118k training images. Each image is annotated with a semantic segmentation mask comprising of 183 classes (inc. an unlabeled class), instance bounding boxes and an image caption. For global caption conditioning, we use the same approach used for class labels in the original EDM2 paper but replace class labels with CLIP captions.
To construct graphs we create two types of nodes: mask nodes and instance nodes.
For mask nodes, we extract the number of unique classes present in the semantic mask. Each mask node is connected to the image via its corresponding class assignment (w.r.t the mask) at the latent resolution. Instance nodes are created per bounding box annotation and connected to pixels within the box. Each instance is connected to the class node of the same type. During training time we randomly drop out ($p=0.5$) mask nodes with uniform probability, ensuring the model does not rely too heavily on mask inputs and must infer the rest of the scene when partial masks are provided.

\textbf{Visual Genome} consists of 108k training images with 3.8M object instances and 2.3M relationships between them. The dataset primarily comprises: object bounding boxes, attributes and relationships. Due to the volume of classes and different relationships between them, we step away from previous image synthesis research using scene graphs, and instead opt for a strategy of encoding classes, attributes and relationships through CLIP \cite{radford2021learningtransferablevisualmodels}, closer inline with prior text-to-image work. For example, to encode the relationship \textit{pulling(horse, carriage)} we extract the latent representation of ``\textit{pulling}'' from CLIP and use this as the directional edge attribute. We apply the same method to attributes via self-loops. Likewise, instance node features are populated in the same fashion via their class label, and are connected to the image w.r.t to their bounding box information. The HIG representation naturally lends itself to complex data and allows a method to account for overlapping regions and relationships to be naturally represented. We filter the attributes/relationships and objects to remove instances that occur less than 250 and 1000 times respectively, in contrast to previous methods that apply stricter thresholds of 500 and 2000.

\textbf{Implementation Details} 
We train models using 4 x A100 GPUs, using training recipes from EDM2, however, due to limited compute resources we reduce the batch-size from $2048$ to $256$ and half the learning-rate from the proposed values, although not optimal we find this to be quicker to train in practice. We train our medium and XXL base models for a combined total of 12 A100 GPU days each, yet do not witness convergence within this timeframe. Our medium model is comparable, in terms of trainable parameters, with previous SOTA, while the XXL is trained to demonstrate scalability. See Appendix Table \ref{table:model_hparams} for full details.

\subsection{Evaluation Metrics.}
We assess quality, diversity and controllability.

\textbf{Fréchet inception distance} (FID) is the primary evaluation metric for visual quality \cite{heusel2018ganstrainedtimescaleupdate}, measuring the distance between feature distributions of generated and real images. It computes the Fréchet distance using features from Inception-v3 \cite{szegedy2015rethinkinginceptionarchitecturecomputer}  with lower values indicating higher quality and realism. We also report $\text{FID}_{\textit{DINOv2}}$ \cite{oquab_dinov2_2024} which has been observed to align better with human preferences \cite{stein_exposing_2023}.

\textbf{Diversity Score} (DS), as introduced in \cite{zheng_layoutdiffusion_2024}, quantifies the perceptual similarity between two images of the same layout generated using different seeds. Specifically, it leverages the learned perceptual image patch similarity (LPIPS) metric \cite{zhang2018perceptual}. A higher LPIPS value corresponds to greater dissimilarity, meaning that a higher DS is desirable for our application.


\textbf{YOLOScore} \cite{LAMA} is used to assess the compliance of the generated samples against prior conditions. To do this, a YOLOv4 \cite{bochkovskiy2020yolov4optimalspeedaccuracy} is used to predict the bounding boxes for a generated sample which are then compared against the ground truth ones, previously used in the HIG for generation. YOLOScore is derived from a combination of several metrics,  including intersection over union, classification accuracy with respect to categorical labels, and confidence scores.

Our metrics are reported on the same validation splits as previous work (5K samples for COCO, 11K samples for VG), however prior work has excessively filtered the sets to improve performance, a practice we do not adopt. FID is calculated between ground-truth validation splits and $\times5$ generations given identical prior conditions. Reported DS values are calculated as the average between all available pair-wise comparisons, and error bars represent the standard deviation between them. Finally, YOLOScore is reported as the average mean precision (\%) of the scores calculated from single corresponding generations. For generations we use a default auto-guidance strength of 1.8, closely in line with the optimal value in EDM2 for $\text{FID}_{\textit{DINOv2}}$, however, as noted in their work, FID considers this optimal value to be a poor choice, and vice versa - see Appendix \ref{fig:fid_appendix}.

\subsection{Quantitative Results}

Quantitative comparisons of our work (at $512\times512$) and the previous SOTA (at $256\times256$) are shown in Table \ref{table:metrics}. For a fair comparison we present results on generated samples with and without mask inputs. We achieve superior performance in all metrics and experiments, despite adopting a less restrictive filtering strategy on validation data. 

\begin{figure}
    \centering
    \includegraphics[width=1\linewidth]{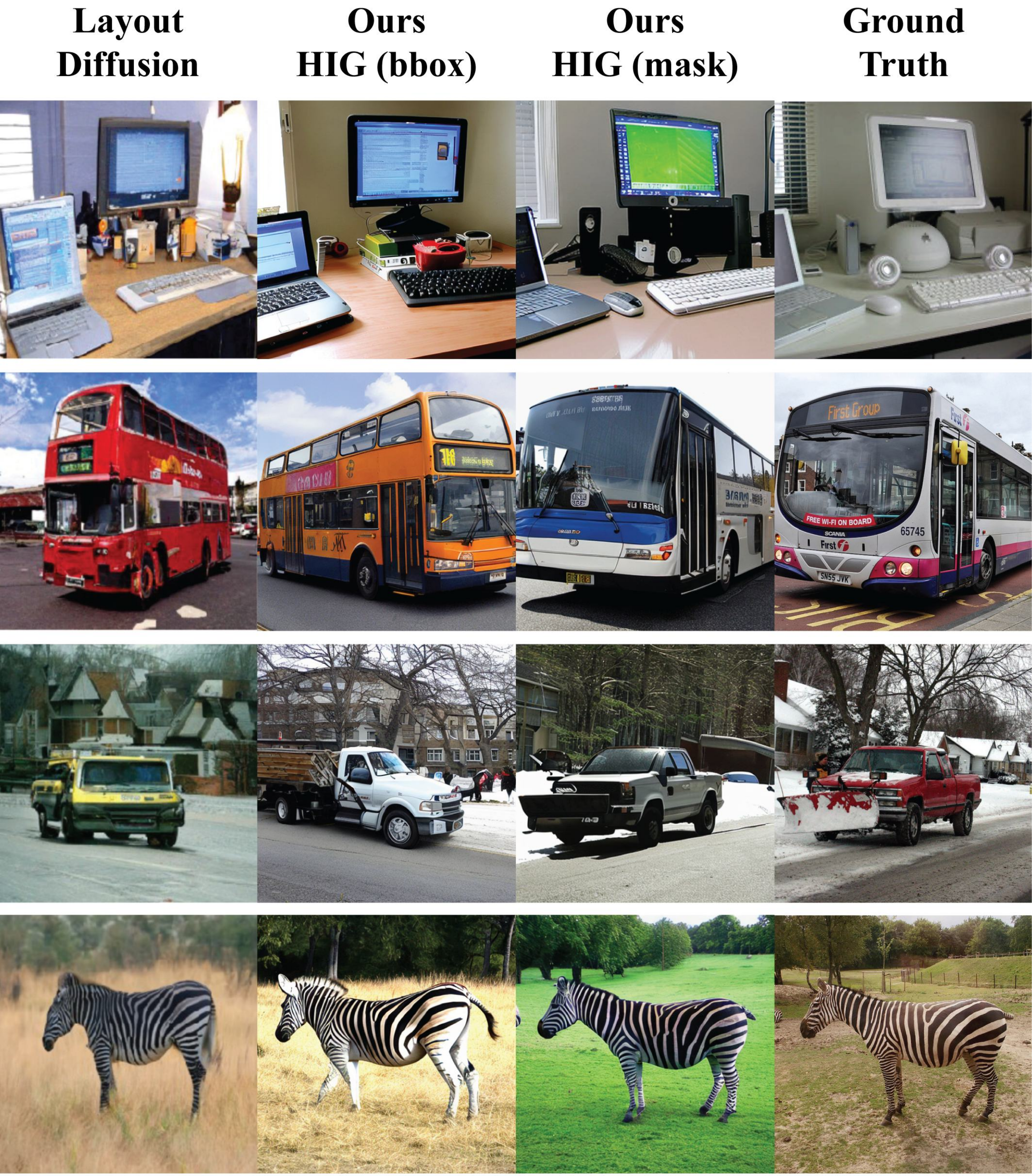}
    \vspace{-20pt}
    \caption{Comparison with SOTA and reference on COCO-stuff.}
    \label{fig:3}    
\end{figure}

\begin{figure}[t]
    \centering
    \includegraphics[width=1\linewidth]{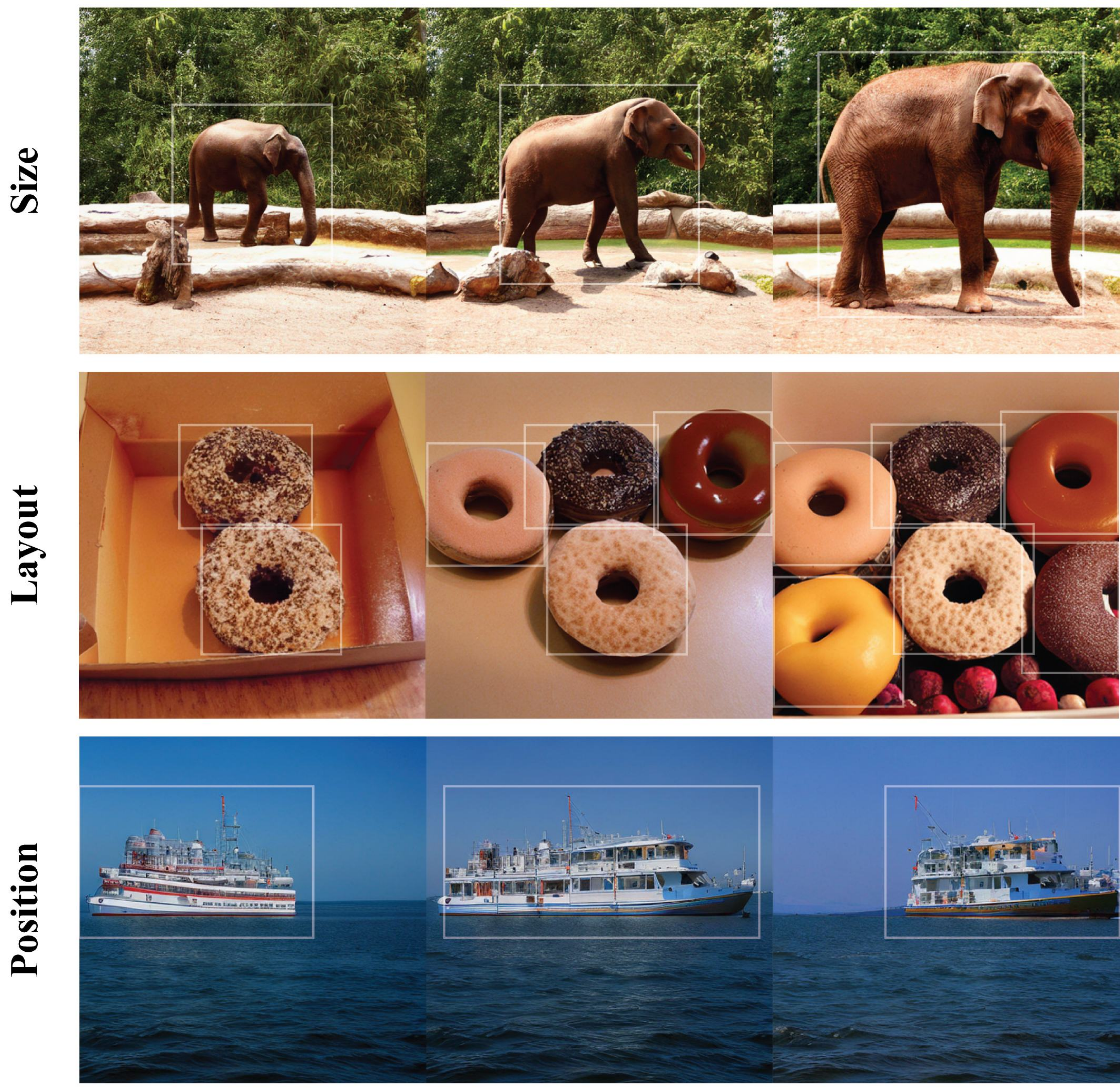}
    \vspace{-20pt}
    \caption{Generations from the same seed demonstrate HIG's ability to control size, quantity and position of objects.}
    \label{fig:size_editing}    
\end{figure}

Across our presented models there are interesting disparities. We see much higher DS for our bounding box models when compared to our mask models - we interpret this as the increase in spatial control decreases the diversity of the generated images. For YOLOScore, we find the performance improves when the mask input is available, likely due to less ambiguity in the conditioning and overlapping bounding boxes taking less effect. In essence, the outline of the object is given for free which likely improves classification accuracy. We see that our XXL model performs worse on this task, suggesting that our larger models may
be less controllable. Given compute constraints, the XXL model remains under-trained and did not plateau during training (notably, the original model was pre-trained for 291 GPU days, whereas ours was trained for only 12.) Perhaps given more training time this scenario would change.

Finally, we observe interesting trends in the FID scores. We notice that standard FID prefers our smaller model, whereas $\text{FID}_{\textit{DINOv2}}$ has a much clearer trend and greatly prefers our XXL model. While standard FID is more sensitive to low-level distributional shifts, $\text{FID}_{\textit{DINOv2}}$ better captures high-level semantic consistency, as detailed in the original DINO work \cite{szegedy2015rethinkinginceptionarchitecturecomputer}. Thus, this preference suggests that our XXL model produces images more close to the ground truth in terms of semantic coherence and object fidelity. This is further supported from visual inspection in the head-to-head comparison shown in Appendix Figure \ref{fig:m_vs_xl_comparison}, where high $\text{FID}_{\textit{DINOv2}}$ seems to align more closely with human-perceived quality.

\subsection{Qualitative Results} 

Our comparison with the current state-of-the-art method \cite{zheng_layoutdiffusion_2024}, illustrated in Figure \ref{fig:3}, demonstrates significant improvements in image quality achieved by our approach. Specifically, HIG with bounding boxes generates highly realistic images, outperforming previous SOTA in terms of detail and coherence. Furthermore, HIG with masks excels in accurately preserving the original sub-structures of the image, as explicitly guided by the given masks. This capability makes HIG with masks particularly valuable in scenarios where precise spatial control is essential. Refer to Appendix Figure \ref{fig:appendix_best} for a selection of high quality generated outputs, and Appendix Figures \ref{fig:appendix_best_diversity}-\ref{fig:appendix_best_box} for diverse generations that demonstrate the difference between mask and bounding box generation.

\textbf{Layout and attribute modification} involves altering the HIG condition through semantic and/or spatial alterations. To evaluate the controllability of the model, we conduct a series of custom generations to guide the model to a desired layout. In Figure \ref{fig:size_editing} we test the model's capability in adjusting size, layout and position of objects in a scene. To make a fair test we keep the random generation seed across all runs the same, which also leads to the effect that the image remains mostly unchanged, a desirable effect in image editing. The model generation is highly controllable through adjusting the spatial edges on the HIG, in all three tests.

\begin{figure}[t]
    \centering
    \includegraphics[width=1\linewidth]{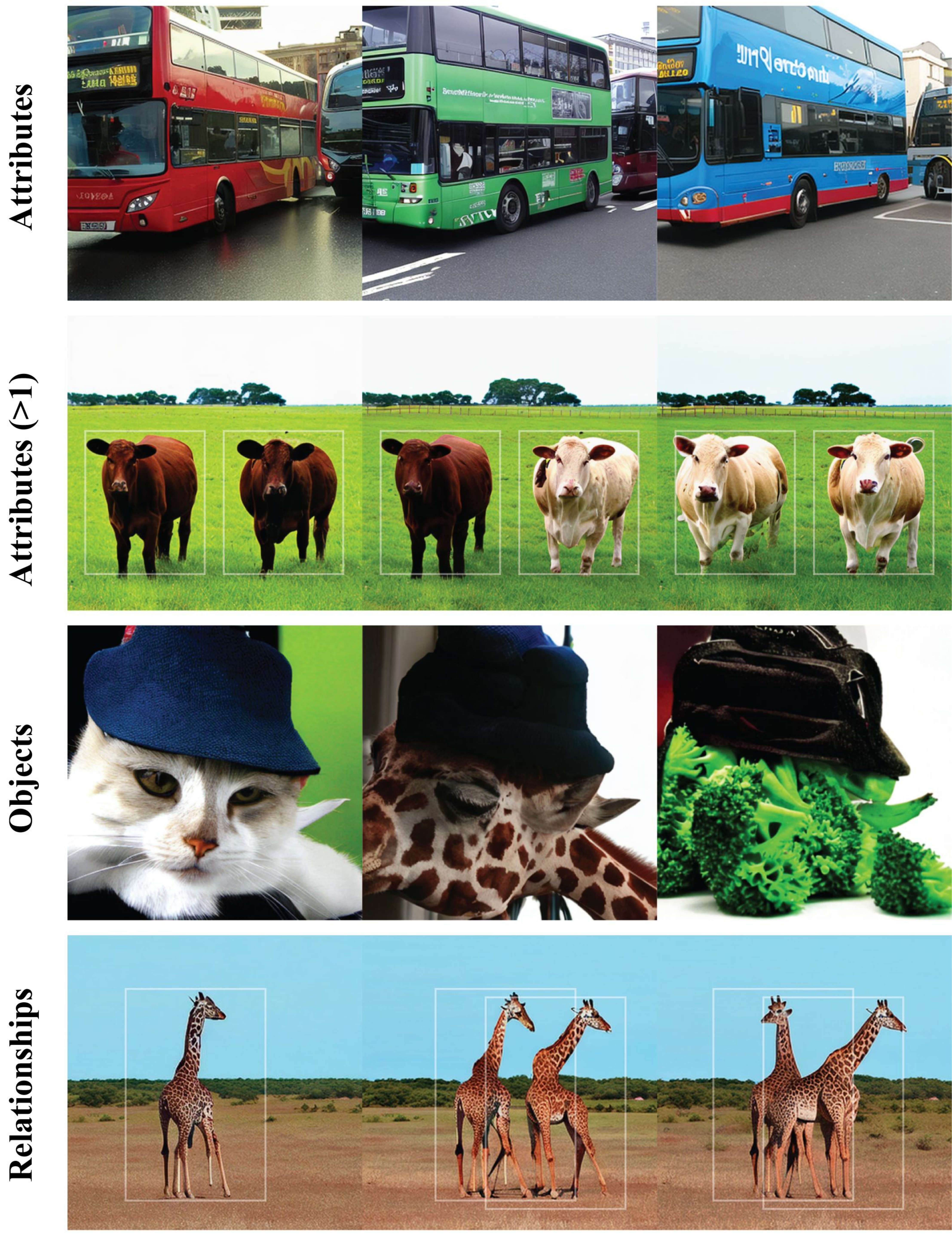}
    \vspace{-20pt}
    \caption{HIG enables precise, localized control over semantic conditions, including attributes, objects, and their relationships.}
    \label{fig:colour_editing}
\end{figure}

\begin{figure}[t]
    \centering
\includegraphics[width=1\linewidth]{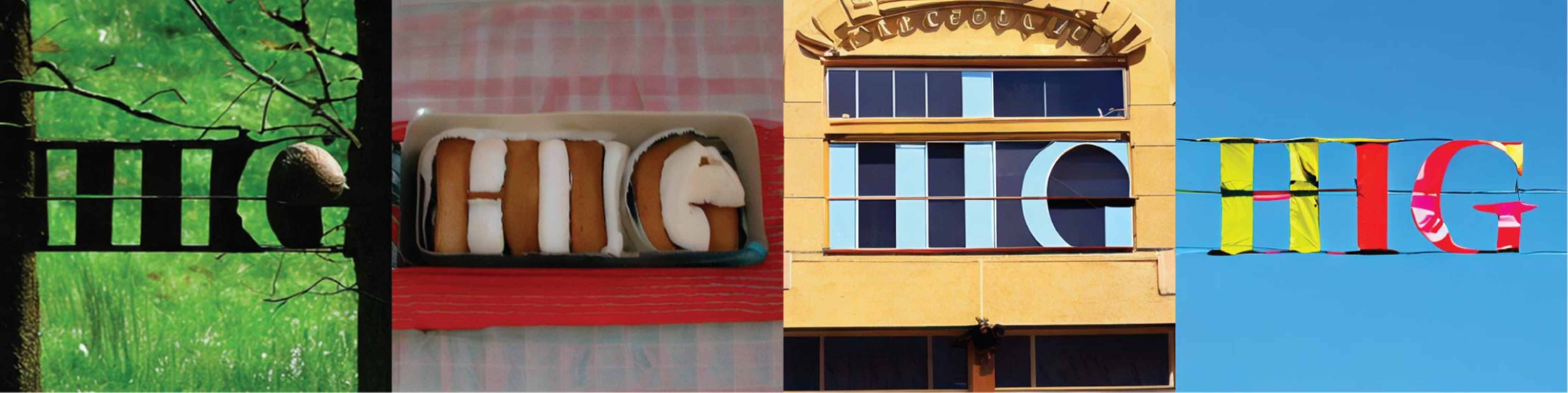}
    \vspace{-20bp}
    \caption{We show HIG can effectively generate consistent images, adhering to underlying mask and semantic conditions. In this case we show `HIG' written in trees, food, windows and kites.}
    \label{fig:fusion}
\end{figure}

We also test the model's capability to adjust semantics in the scene, for example modifying attributes (e.g. colour), objects (e.g. category) and relationships (e.g. behind/in front.) Our results in Figure \ref{fig:colour_editing} show unprecedented control, enabling high-quality image generation with precise localized edits (e.g., changing the color of a specific cow.) Furthermore, our model exhibits strong generalizability by effectively generating out-of-distribution object combinations, such as a cat-shaped broccoli, positioned beneath a hat. While object-dependent spatial masks pose challenges when combined with semantic guidance, the model preserves consistency and quality, further demonstrated in Figure \ref{fig:fusion}. Lastly, the model respects relationships between two ambiguous overlapping bounding boxes (e.g., positioning one giraffe in front of another), establishing a new benchmark for structured and context-aware image generation, see Appendix Figures \ref{fig:relationships appendx}-\ref{fig:elephant_appendix} for more examples.

\subsection{Ablation Experiments}
We conduct a small ablation study on various architecture design choices. We launch smaller training runs to isolate factors of GNN depth, control integration and magnitude preservation. In Table \ref{table:ablation} we show that decreasing the number of HIG blocks degrades performance. We also find that applying magnitude preservation functions to incoming conditioning features is essential for the ControlNet, by only using regular addition (and zero-gain) we find NaNs start to appear in the training loss as feature magnitudes explode. We also investigate whether preserving magnitudes in the GNN is essential or if standard PixNorm suffices. Interestingly, our tests show performance comparable to our base MP-GNN approach. Finally, we experiment whether full-fining is a viable alternative to ControlNet, despite scoring the lowest $\text{FID}_{\textit{DINOv2}}$ we witness poor perceptual quality in comparison to the base model. We leave it to future research to develop upon these results further, for example by training to convergence.
\begin{table}[t]
\centering
\caption{Ablation study results on COCO-stuff at 12M images.} 

\begin{tabular}{@{}lcc@{}}
\toprule

& \textbf{FID ↓} & \textbf{$\text{FID}_{\textit{DINOv2}}$ ↓} \\ \cmidrule{2-3}
HIG-Medium & \textbf{18.34} & 332.00 \\
\midrule 
Depth (2 Layers) & 20.40 & 366.09 \\
Without MP ControlNet & \textit{NaN} & \textit{NaN} \\
Without MP GNN (w/ PixNorm) & 18.56 & 328.67 \\
Full Fine-tuning & 19.66 & \textbf{325.297} \\
\bottomrule
\end{tabular}
\label{table:ablation}
\end{table}

\section{Discussion, Limitations and Future Work}
\label{sec:discussion}

Our novel HIG representation enables highly controllable and complex conditioning, outperforming the state-of-the-art in image quality while remaining more computationally efficient by foregoing quadratic attention mechanisms. By structuring images and conditioning variables as a single heterogeneous graph, we incorporate all available data—including attributes and relationships—into a unified representation. This allows for unprecedented control over local attribute conditioning and complex spatial ambiguities, paving the way for larger and more diverse datasets in image synthesis. While we attempt to mitigate randomness by using consistent seeds across experiments, our model still exhibits notable failures in adhering to the prescribed relationships and attribute conditioning, which we attribute to factors such as inconsistent spatial labeling, under-training, and other contributing limitations.
Despite relying on basic graph convolutional operators, our model achieves strong results. Future work will explore more advanced architectures, such as attention-based message passing or higher-order graph operators, which could further enhance expressivity. Moreover, we only utilise models trained on ImageNet as our backbone, we believe a more generalised model would significantly improve our results. More broadly, we believe our approach has implications beyond diffusion models, offering new directions for structured representations in generative modeling.

\clearpage
\newpage

\section*{Impact Statement}

Large-scale image generation models present notable societal risks, including the spread of disinformation and the reinforcement of stereotypes \cite{eiras2024risksopportunitiesopensourcegenerative}. While the methods we introduce in this paper significantly improve the controllability of these models, they may also intensify these risks by making it easier to generate tailored, high-fidelity content that aligns precisely with specified scenes, faces, and events, potentially amplifying the spread of misleading or biased information.

The substantial computational demands required for training and sampling diffusion models should also be acknowledged, as they lead to considerable energy consumption and may further contribute to broader environmental challenges, including climate change.

\section*{Acknowledgments} We would like to acknowledge support from the Engineering and Physical Sciences Research Council (EPSRC) Ph.D. Studentship EP/N509620/1 and the UKRI access to high performance computing facilities program.

\section*{Conflicts of Interest} SWP is co-founder of a spin-out company called Matta that develops AI for Factories.


\bibliography{icml}
\bibliographystyle{icml2023}

\renewcommand{\thesection}{A\arabic{section}}
\renewcommand{\thesubsection}{A\arabic{section}.\arabic{subsection}}

\appendix
\clearpage
\onecolumn
\setcounter{page}{1}  
\setcounter{section}{0}  
\section*{Appendix}

\section{Sum of Random Unit Vectors}
\label{appendix:sum_random}

First, recall that for random zero-mean vectors, their expected Euclidean norm $\mathbb{E}[\|\mathbf{a}\|^2] = \mathbb{E}[a_1^2 + a_2^2 + \dots + a_d^2] $ is related to the variance, since $
\text{Var}(a_i) = \mathbb{E}[a_i^2] - (\mathbb{E}[a_i])^2 = \mathbb{E}[a_i^2]$. Due to the independence assumptions, we can rewrite the expected norm in terms of the variance as $\mathbb{E}[\|\mathbf{a}\|^2] = \mathbb{E}[a_1^2] + \mathbb{E}[a_2^2] + \dots + \mathbb{E}[a_d^2].$

Let us consider the sum of \(N\) independent random \textit{unit} vectors \(\mathbf{c} = \sum_{i=1}^{N} \mathbf{a}_i\), where due to independence zero-mean expectation \(\mathbb{E}[\mathbf{a}_i \mathbf{a}_j] = \mathbb{E}[\mathbf{a}_i]\mathbb{E}[\mathbf{a}_j] = 0\) for \(i \neq j\). Then,

\begin{equation}
\mathbb{E}[\|\mathbf{c}\|^2] = \frac{1}{N_c} \sum_{i=1}^{N_c} \mathbb{E} \left[ \left( \sum_{i=1}^{N}  \mathbf{a}_i \right)^2 \right]
\end{equation}

\begin{equation}
= \frac{1}{N_c} \sum_{i=1}^{N_c} \mathbb{E} \left[ \sum_{i=1}^{N} \mathbf{a}_i^2 + 2 \sum_{i=1}^{N}\sum^{N}_{\substack{j=1 \\ j < i}} \mathbf{a}_i \mathbf{a}_j \right]
\end{equation}

\begin{equation}
= \frac{1}{N_c} \sum_{i=1}^{N_c} \left[ \sum_{i=1}^{N}  \mathbb{E}[\mathbf{a}_i^2] 
+ 2 \sum^{N}_{i=1}\sum^{N}_{\substack{j=1 \\ j < i}} \underbrace{\mathbb{E}[\mathbf{a}_i \mathbf{a}_j]}_{= 0} \right]
\end{equation}

\begin{equation}
= \sum_{i=1}^{N} \mathbb{E}[\mathbf{a}_i^2].
\end{equation}

If the inputs are standardised, this further simplifies to:

\begin{equation}
\mathbb{E}[\|\mathbf{c}\|] = \sqrt{\sum_{i=1}^{N} \mathbb{E}[\mathbf{a}_i^2]} = \sqrt{N}.
\end{equation}

A normalised version of \(\mathbf{c}\) is therefore: 

\begin{equation}
\hat{\mathbf{c}} = \frac{\mathbf{c}}{\mathbb{E}[\|\mathbf{c}\|]} = \frac{\sum_{i=1}^{N} \mathbf{a}_i}{\sqrt{N}}.
\label{eq:normalised_c}
\end{equation}

We use this simple magnitude preserving summation formula for the aggregations across both meta paths ($\Phi$) and for our neighbourhood pooling operation. Since this is trivial to compute on the fly, this allows variable size neighbourhoods to be included into our operator in Equation \ref{eq:hignn_operator} without increasing its magnitude.

\section{EDM2 Mathematical Preliminaries}
\label{appendix:edm2_preliminaries}

We apply a brief summary of relevant equations for reader convenience. For a complete overview, readers should reference \cite{karras_analyzing_2024} and their exhaustive appendix.

We use forced weight magnitude preservation, as per their Equation 47:

\begin{equation}
\hat{w_i} = \frac{w_i}{\|w\|_2 + \epsilon}.
 \label{eq:karras_47}
\end{equation}

Where $\epsilon$ is a small delta to avoid numerical issues. Readers should reference Karras et al. Algorithm 1 for an implementation of forced weight normalisation. Additionally, we utilise magnitude preserving sum (their Equation 88) at several locations in the network, including integration with the control net:

\begin{equation}
\text{MP-Sum}(\mathbf{a}, \mathbf{b}, t) = \frac{(1 - t) \mathbf{a} + t \mathbf{b}}{\sqrt{(1 - t)^2 + t^2}},
 \label{eq:mp_sum}
\end{equation}

and magnitude preserving concatenation (their Equation 103):

\begin{equation}
\text{MP-Cat}(\mathbf{a}, \mathbf{b}, t) = \sqrt{\frac{N_a + N_b}{(1 - t)^2 + t^2}} \cdot \left[ \frac{1 - t}{\sqrt{N_a}} \mathbf{a} \oplus \frac{t}{\sqrt{N_b}} \mathbf{b} \right].
 \label{eq:mp_cat}
\end{equation}

\section{Implementation parameters}
\label{appendix:parameter_table}


\begin{table}[h]
\centering
\caption{Training details for HIGnn diffusion models.}
\begin{tabular}{@{}lcc@{}}
\toprule
\textbf{Model details}                      &\textbf{ M} & \textbf{XXL} \\ \midrule
Number of GPUs                     &  4 &  4   \\
Minibatch size                     &  256 & 256    \\
Duration                           & 40M &  15M   \\
Channel multiplier                 & 256 & 448    \\
Number of HIG blocks  &  4 &  4    \\
Dropout probability                & 10\% &  10\%   \\
Learning rate max ($\alpha_{ref}$)   & 0.0045 & 0.003    \\
Learning rate decay ($t_{ref}$)      & 10000 & 10000   \\
Noise distribution mean ($P_{mean}$) &  -0.4 &  -0.4   \\
Noise distribution std. ($P_{std}$)  &  1.0 &  1.0    \\ 
Base Model capacity (Mparams)  &  497.8 &  1523.2    \\
ControlNet capacity (Mparams)  &  163.2 &  495.3   \\
HIGnn capacity (Mparams)  &  7.0 &  18.9    \\
Total capacity (Mparams)  &  668.0 &  2037.5    \\
Training time (days) &  3.0 &  3.0    \\
EMA &  None &  None    \\
\bottomrule
\end{tabular}
\label{table:model_hparams}
\end{table}

\newpage

\begin{figure}
    \centering
    \includegraphics[width=1\linewidth]{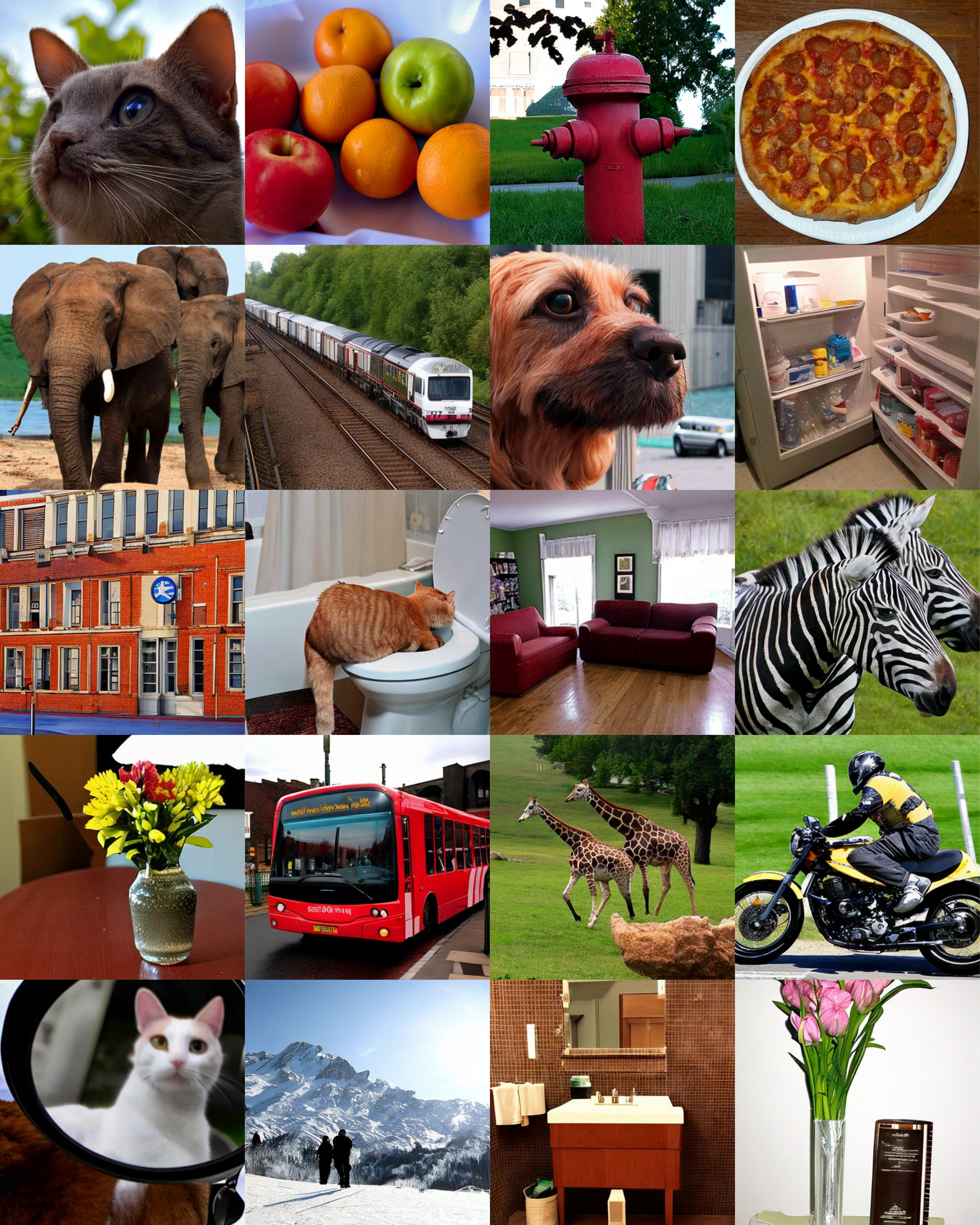}
    \caption{Selected generated samples from HIG using COCO validation conditions.}
    \label{fig:appendix_best}
\end{figure}

\begin{figure}
    \centering
    \includegraphics[width=0.5\linewidth]{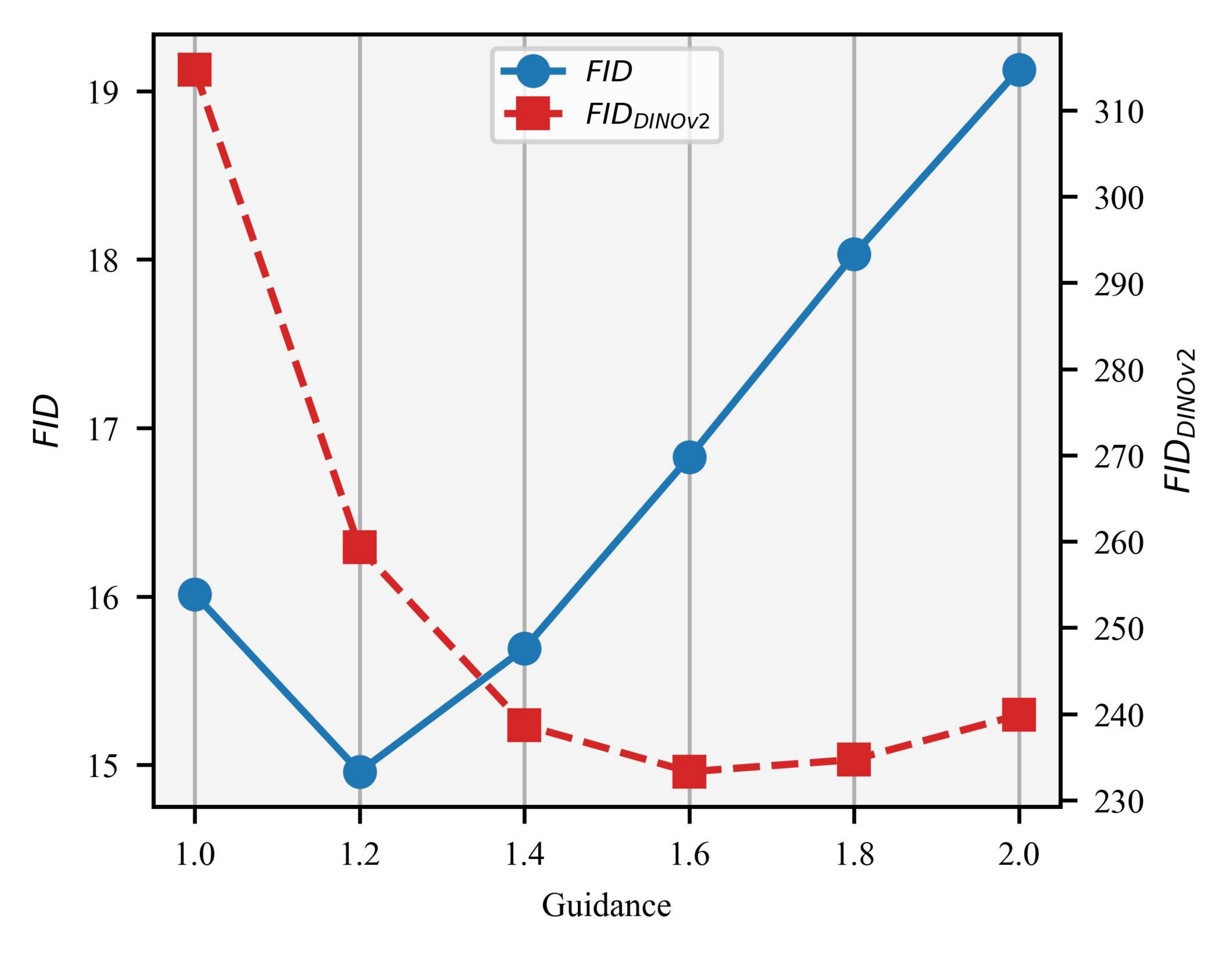}
    \caption{\text{FID} and $\text{FID}_{\textit{DINOv2}}$ vs guidance. To understand the relationship between auto-guidance strength and FID values we generate 5K images for coco validation (i.e. one seed). The relationship we witness is similiar to that reported in the original \cite{karras_analyzing_2024} work, they report optimal FID at 1.4, and optimal $\text{FID}_{\textit{DINOv2}}$ at 1.9. Our optimal values are both slightly lower than this.}
    \label{fig:fid_appendix}
\end{figure}

\newpage

\begin{figure}
    \centering
    \includegraphics[width=1\linewidth]{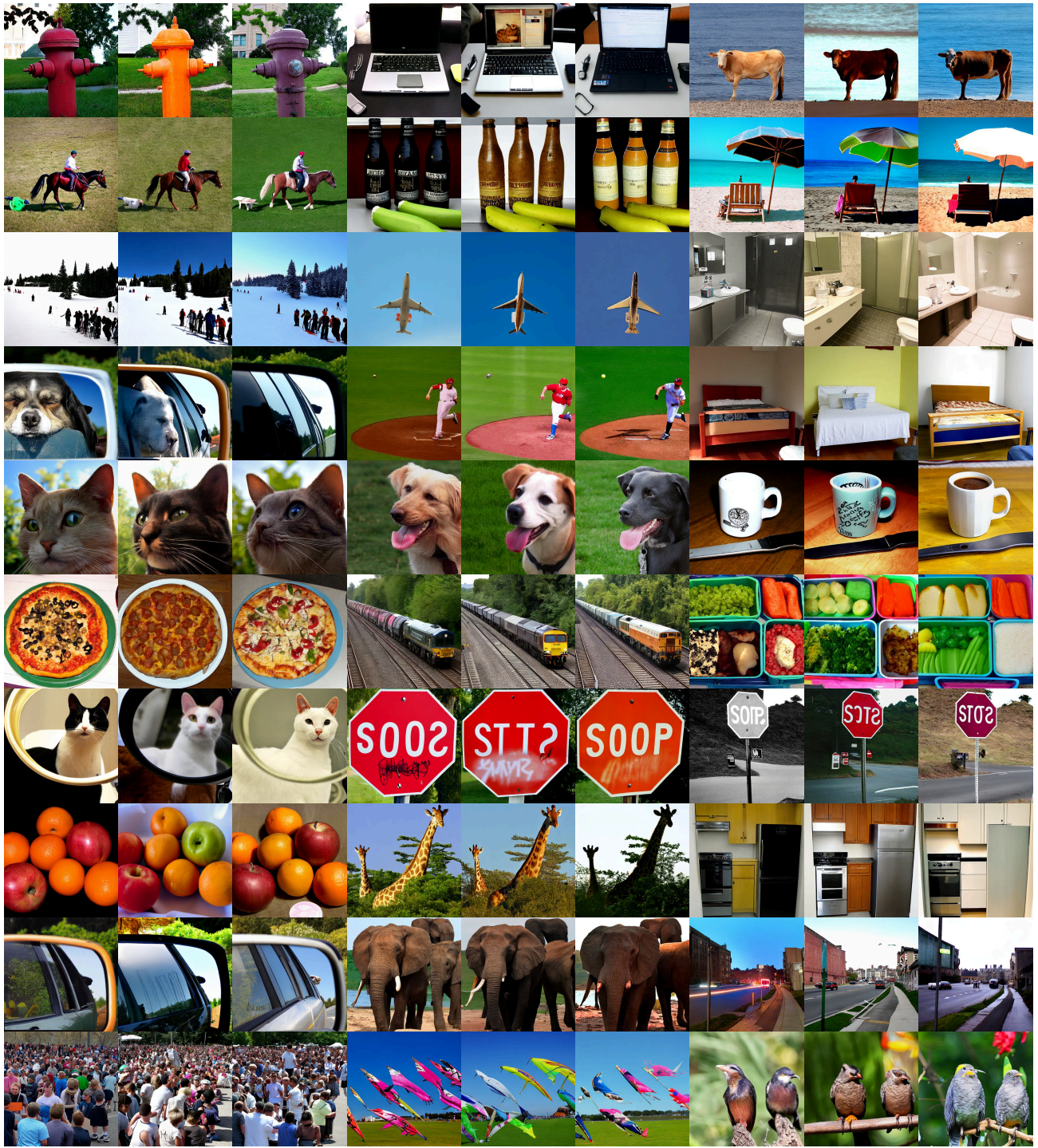}
    \caption{Diverse HIG samples using mask and bounding boxes from COCO validation set.}
    \label{fig:appendix_best_diversity}
\end{figure}

\newpage

\begin{figure}
    \centering
    \includegraphics[width=1\linewidth]{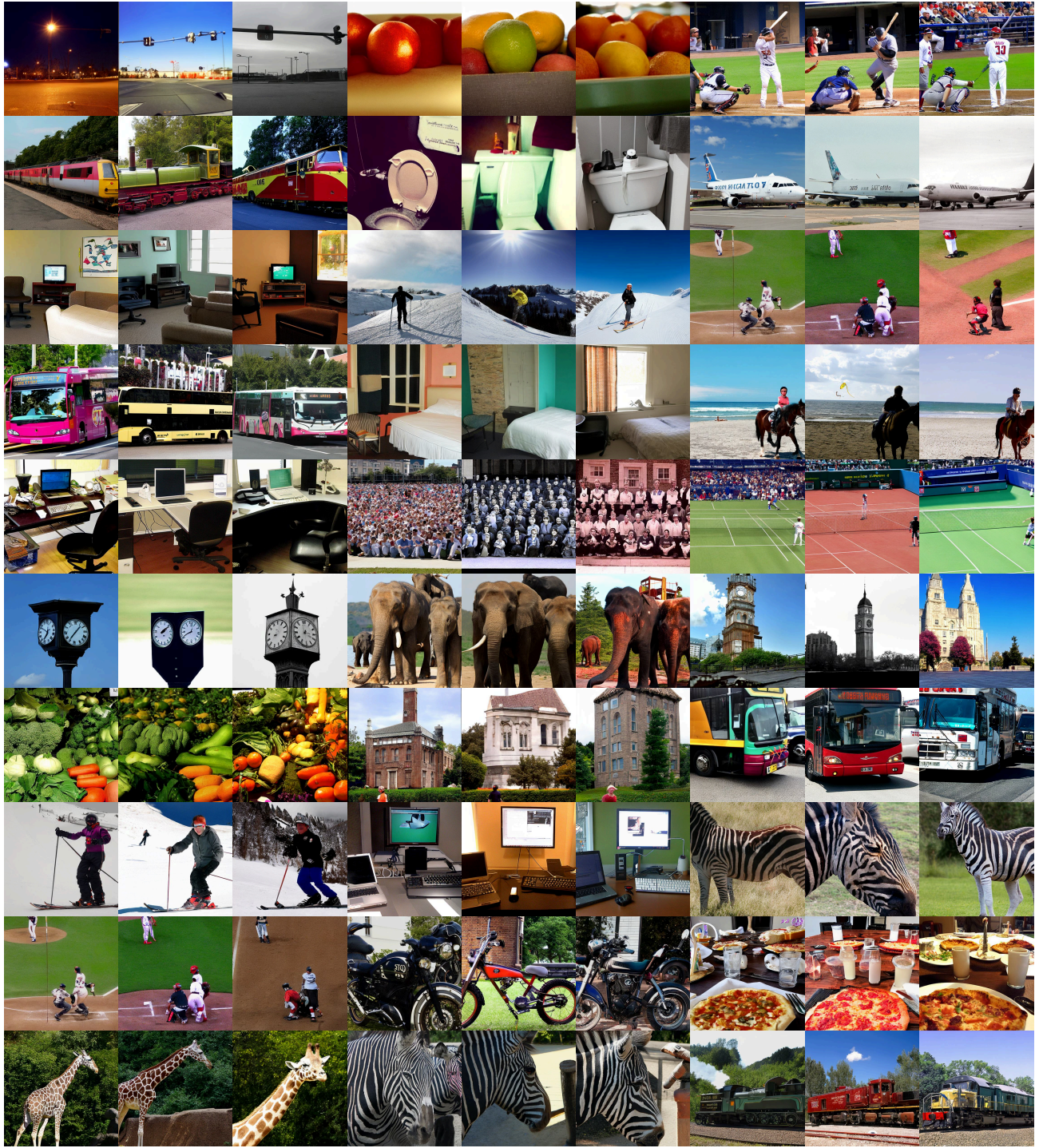}
    \caption{Diverse HIG samples using only bounding boxes from the COCO validation set. We naturally observe increased diversity in images when compared to Figure \ref{fig:appendix_best_diversity} that uses mask inputs.}
    \label{fig:appendix_best_box}
\end{figure}

\clearpage
\newpage

\begin{figure*}
    \centering
    \includegraphics[width=1\linewidth]{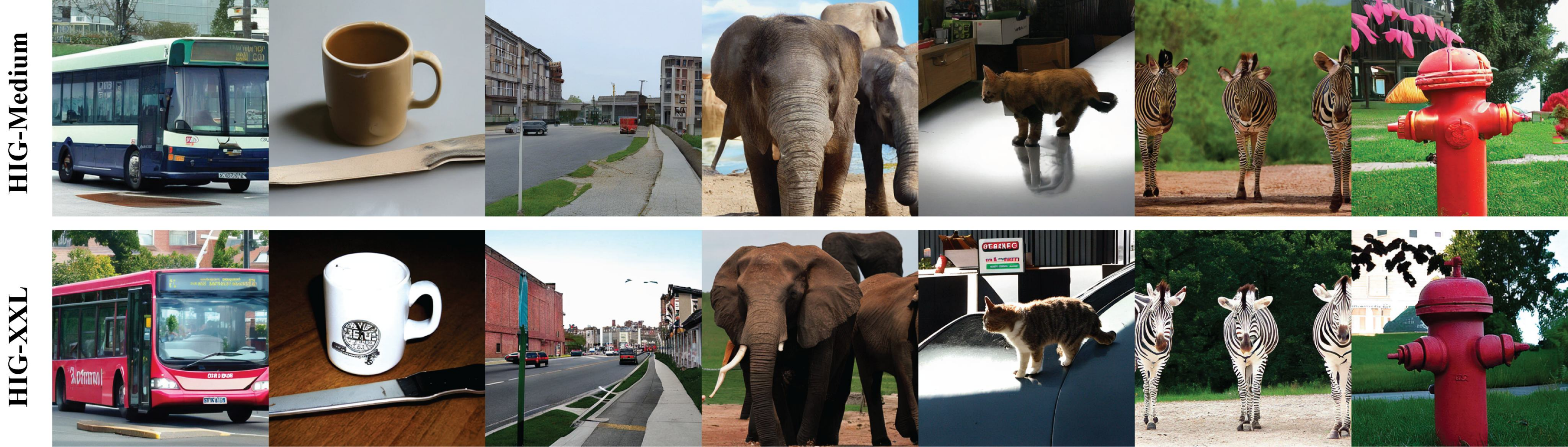}
    \caption{Head-to-head comparisons between generations from HIG-Medium and HIG-XXL. There is a notable enhancement in realism in the latter, which exhibit richer color depth and more convincing surface textures.}
    \label{fig:m_vs_xl_comparison}
\end{figure*}

\begin{figure*}
    \centering
    \includegraphics[width=1\linewidth]{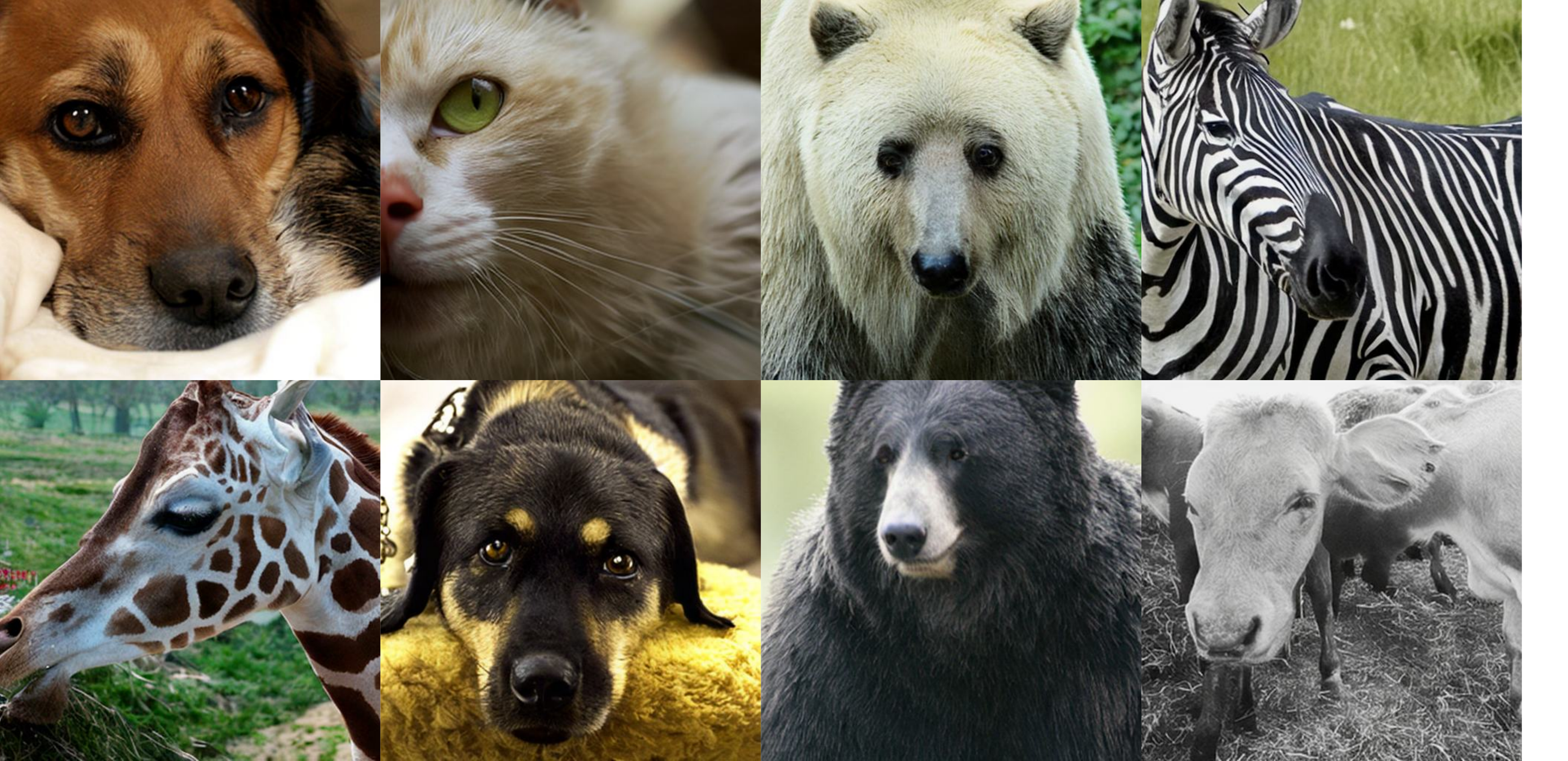}
    \caption{Selected generations using a single bounding box with an animal class label.}
    \label{fig:animal_appendix}
\end{figure*}

\clearpage
\newpage

\begin{figure*}
    \centering
    \includegraphics[width=1\linewidth]{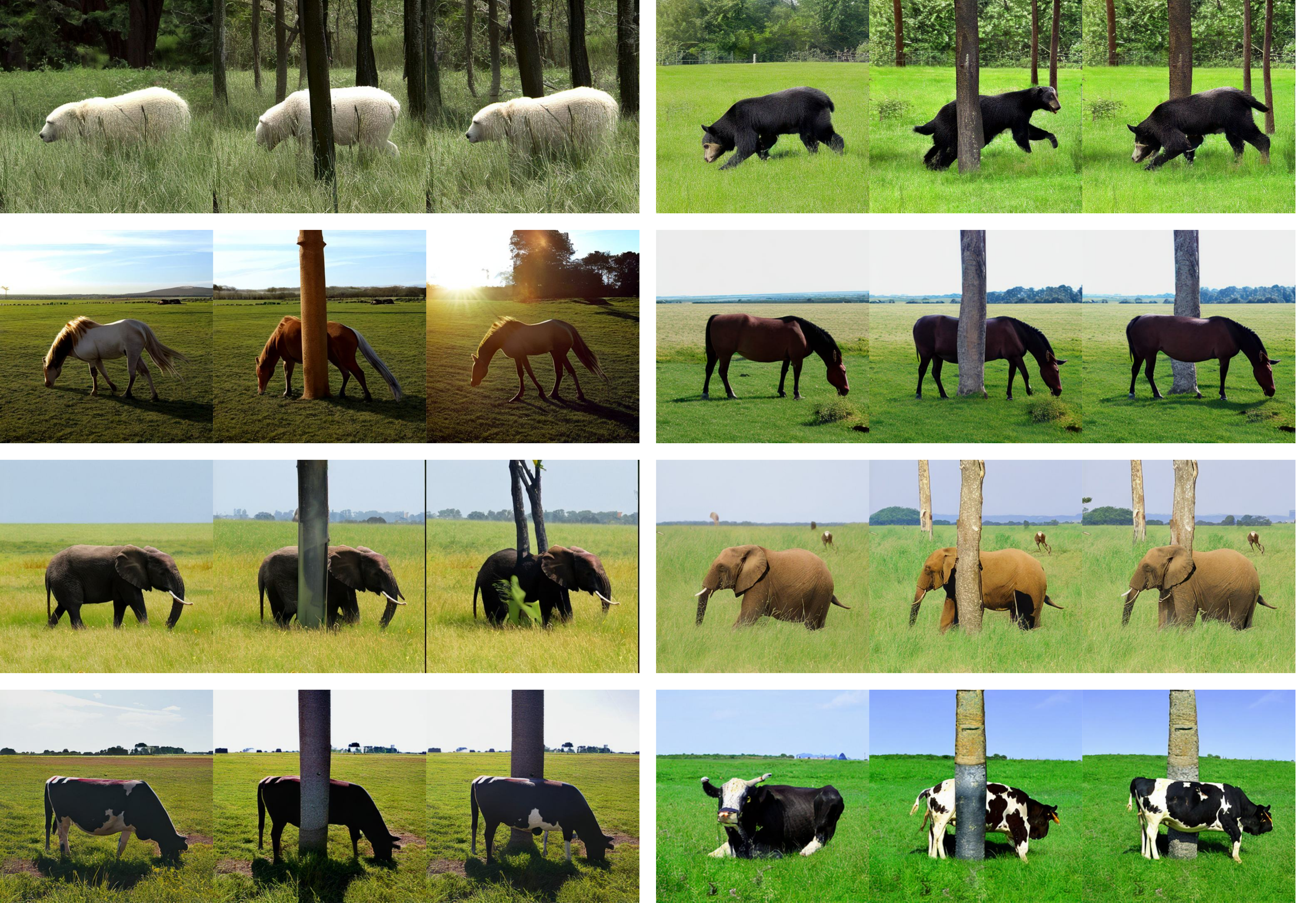}
    \caption{We showcase the ability of the HIG to disambiguate overlapping bounding boxes. In this figure we show bear, horse, elephant and cow bounding boxes overlapped with a tree bounding box,  with the relationship `in front' or 'behind'. We also show the bounding box when the tree is removed.}
    \label{fig:relationships appendx}
\end{figure*}   

\begin{figure*}
    \centering
    \includegraphics[width=1\linewidth]{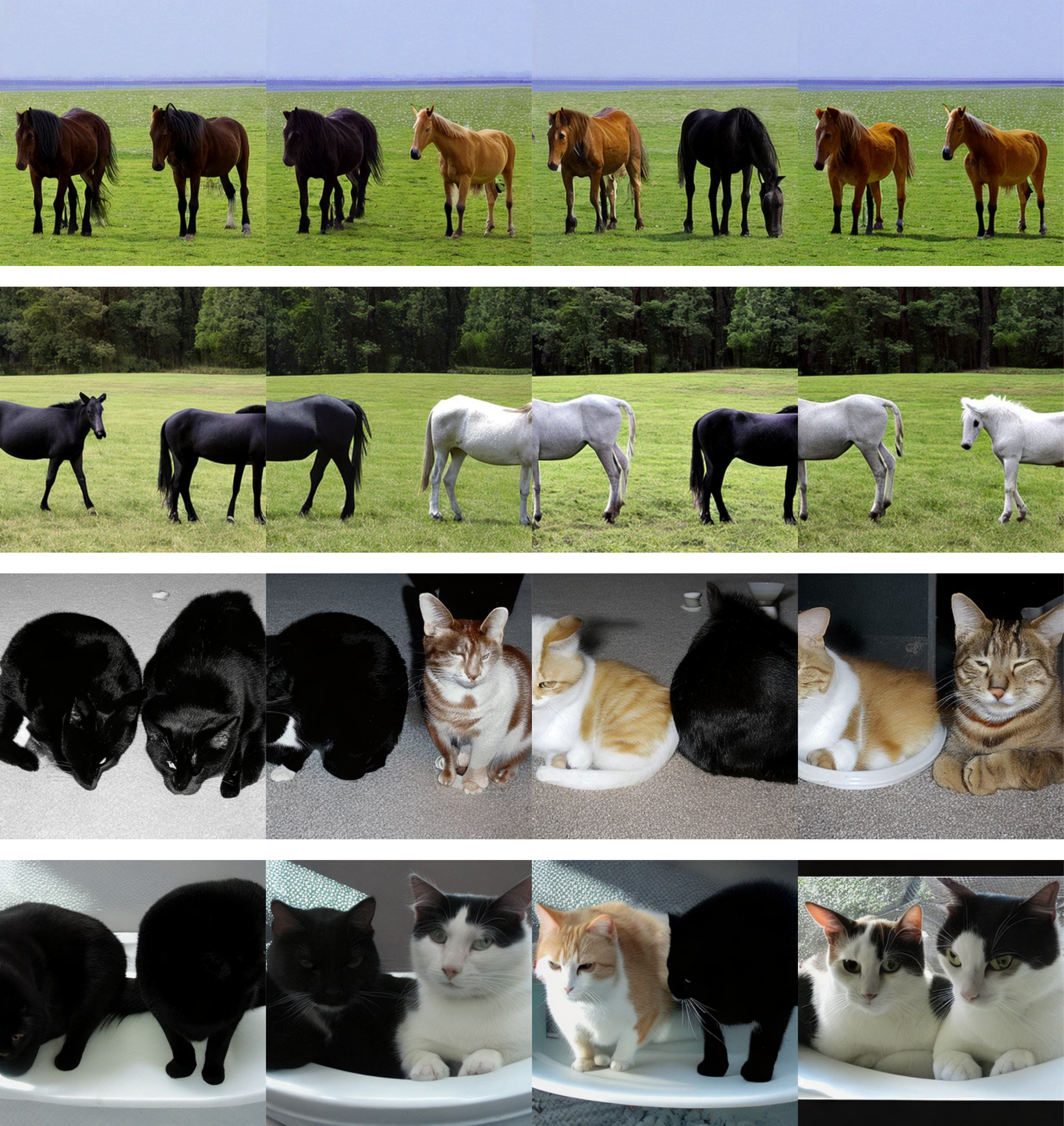}
    \caption{We showcase the ability to locally edit the attribute of an object i.e. the colour of a specific animal in an image.}
    \label{fig:animal_color_appendix}
\end{figure*}

\begin{figure*}
    \centering
    \includegraphics[width=1\linewidth]{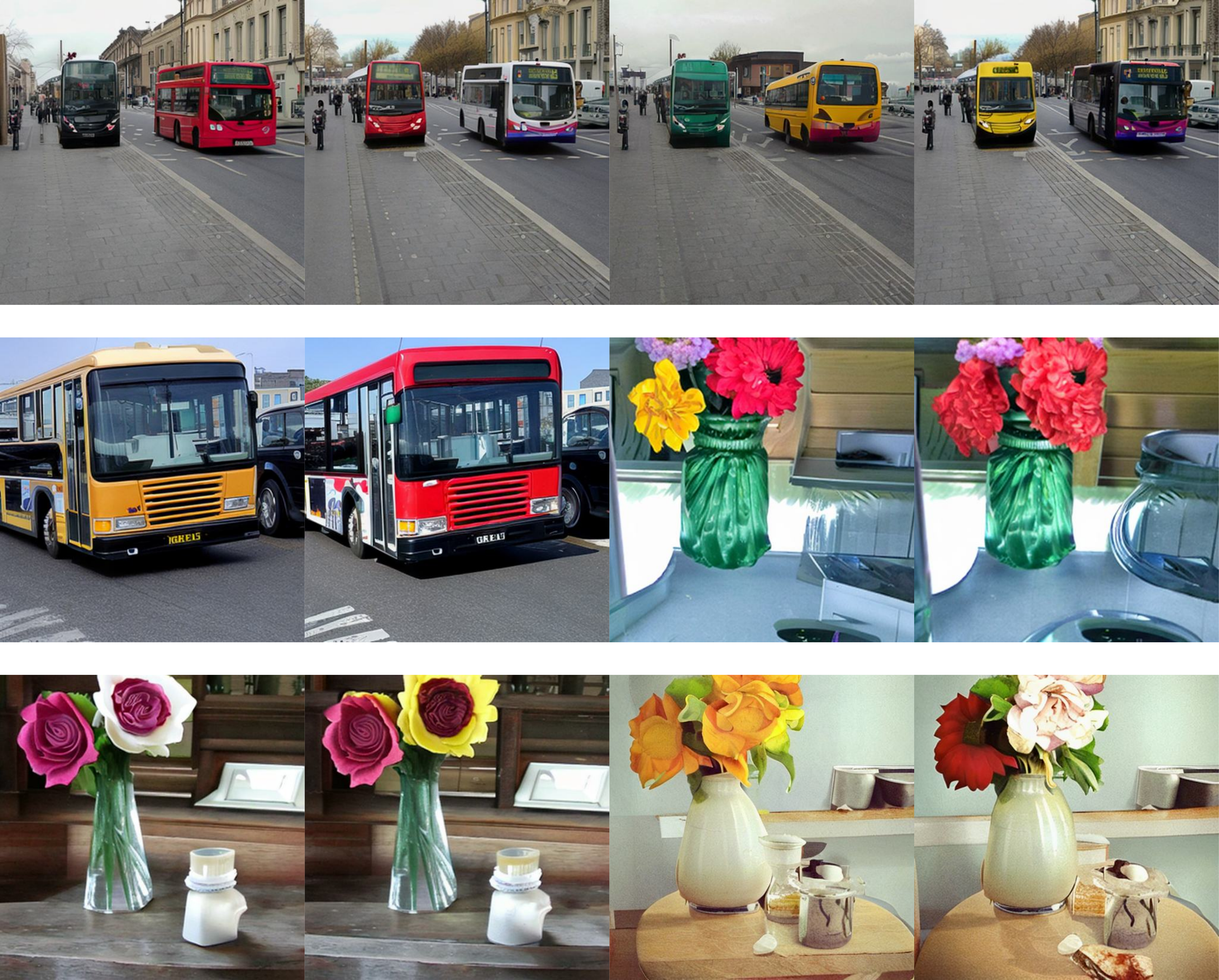}
    \caption{We showcase the ability to locally edit the attribute of an object i.e. the colour of a specific bus or flower in a generation.}
    \label{fig:color_appendix_1}
\end{figure*}   

\begin{figure*}
    \centering
    \includegraphics[width=1\linewidth]{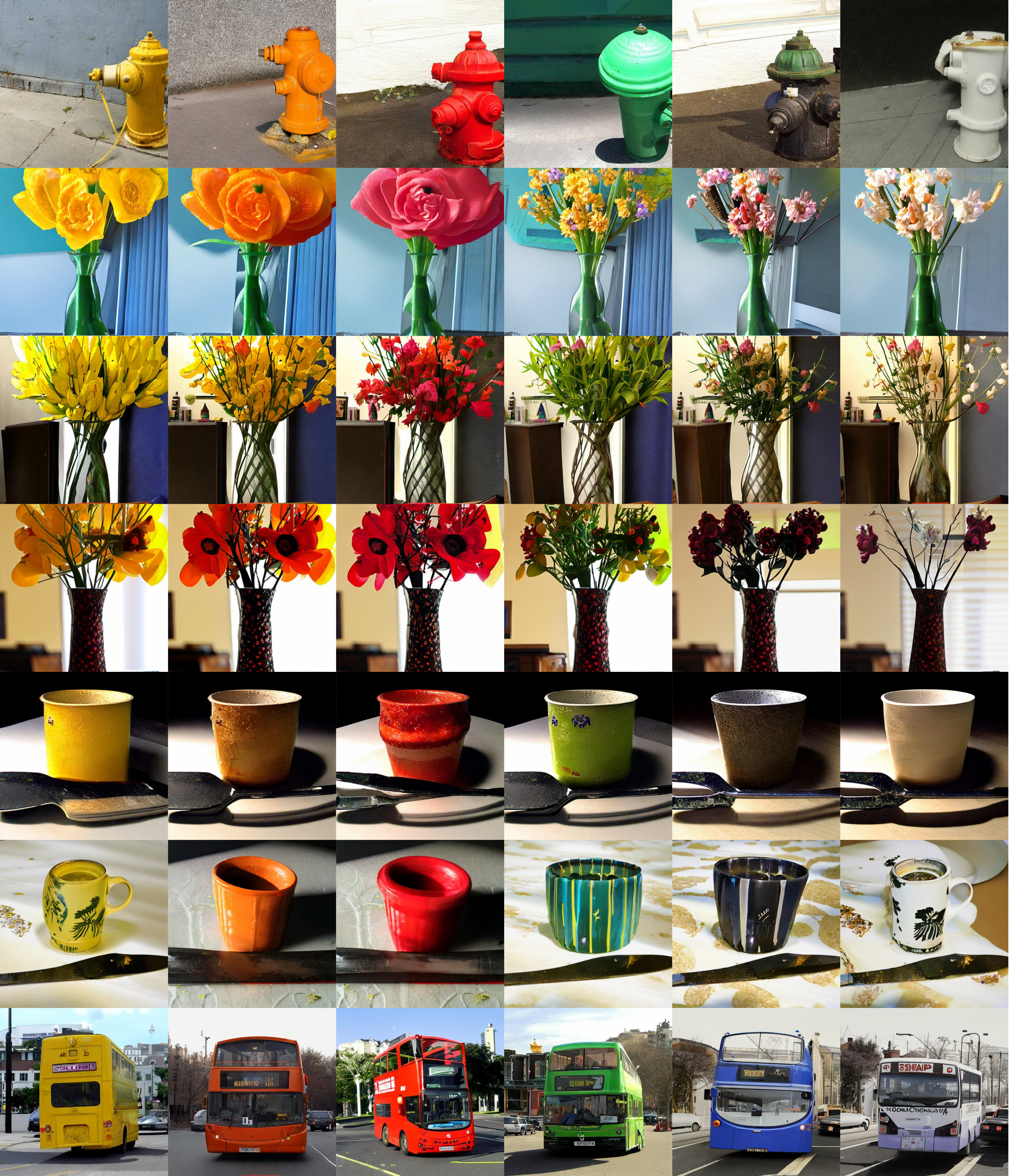}
    \caption{Generations with different colour attributes. }
    \label{fig:color_appendix_2}
\end{figure*}   

\begin{figure*}
    \centering
    \includegraphics[width=1\linewidth]{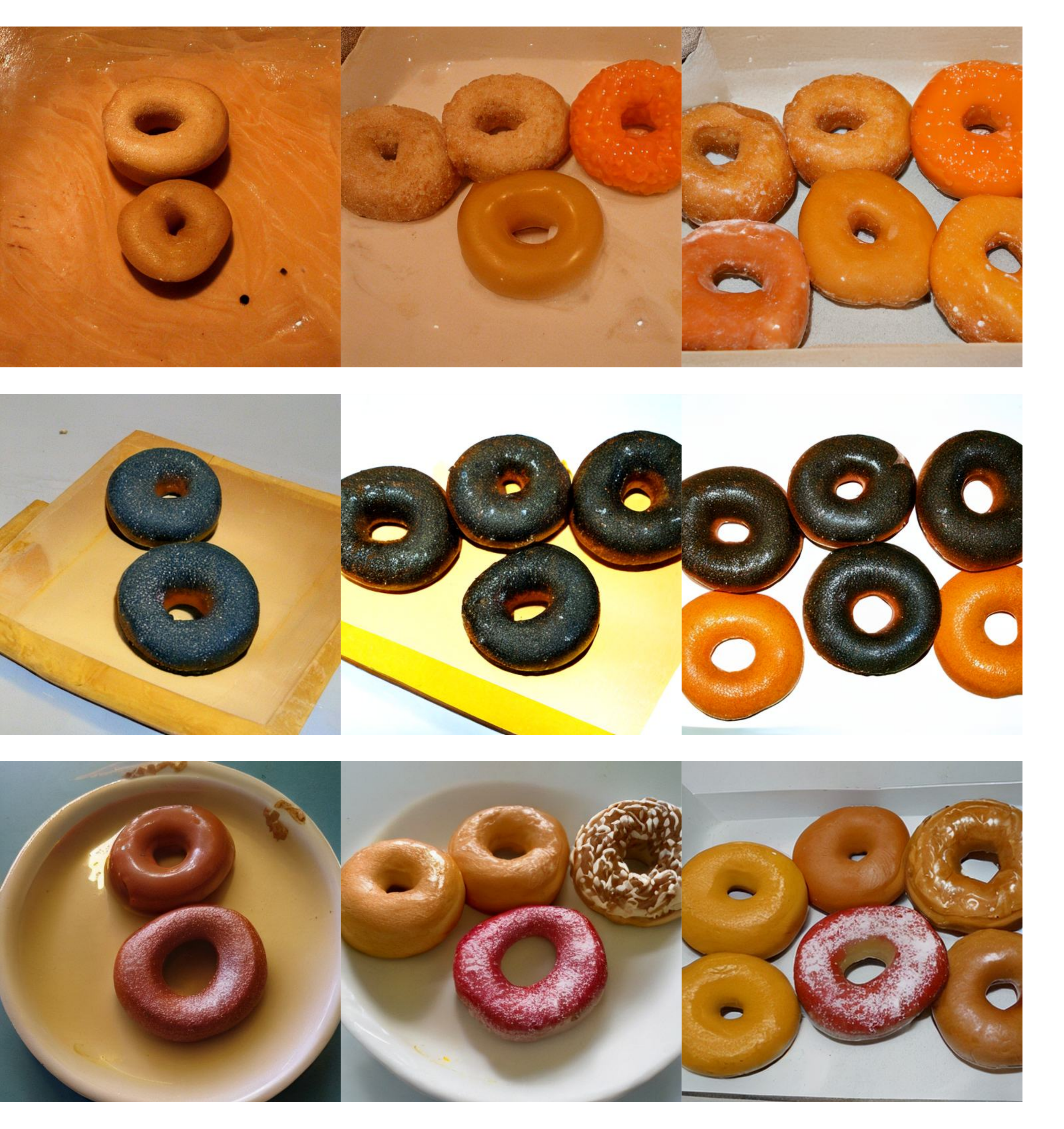}
    \caption{Editing a scenes layout. We observe that when generating images with the same seed that object semantics stay relatively consistent as we add or remove objects from a scene. }
    \label{fig:donut_appendix}
\end{figure*}   

\begin{figure*}
    \centering
    \includegraphics[width=1\linewidth]{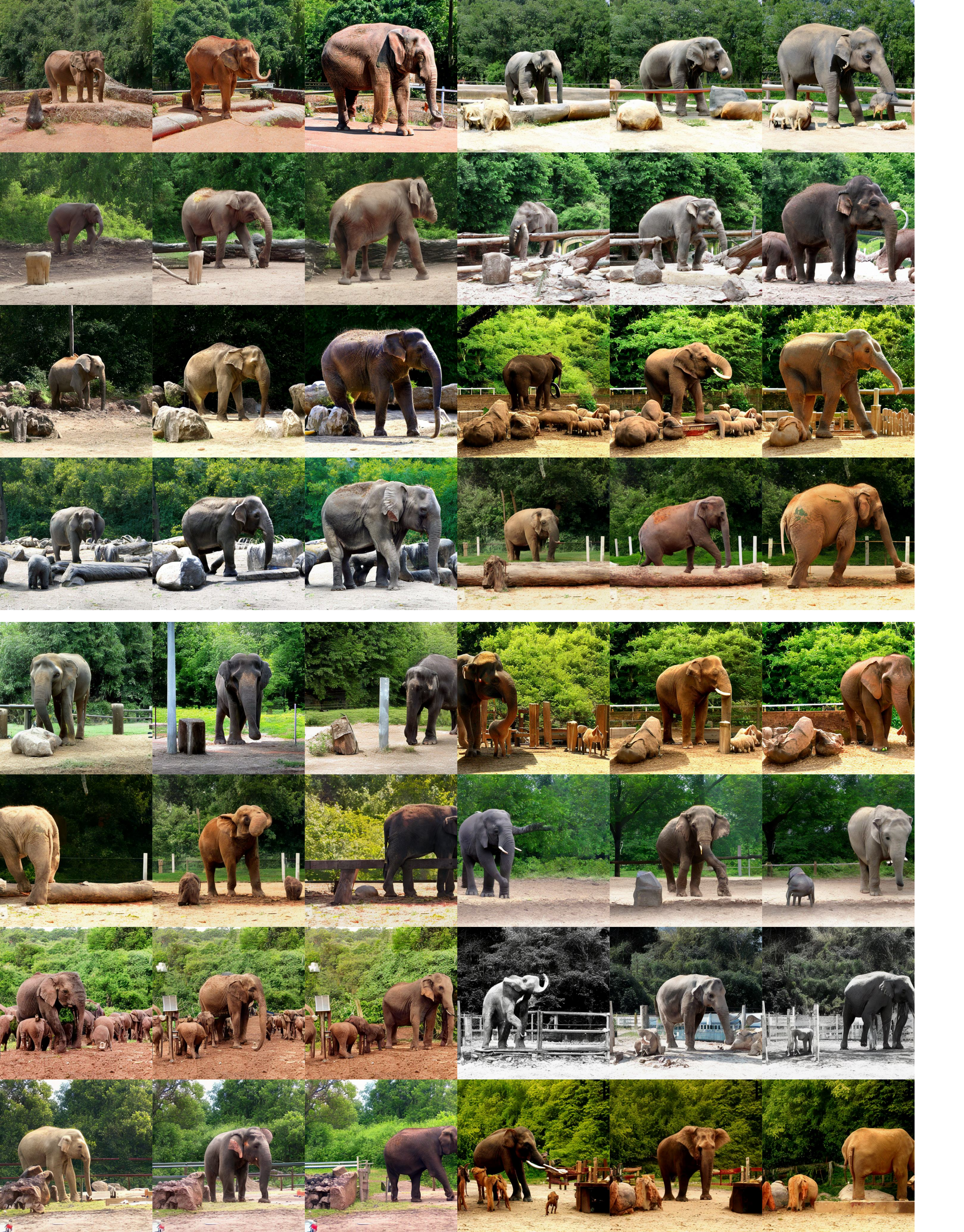}
    \caption{We showcase the ability to change the size and position of an object in a scene.}
    \label{fig:elephant_appendix}
\end{figure*}

\end{document}